\documentclass{article}

\usepackage{arxiv}

\usepackage[utf8]{inputenc} 
\usepackage[T1]{fontenc}    
\usepackage{hyperref}       
\usepackage{url}            
\usepackage{booktabs}       
\usepackage{amsfonts}       
\usepackage{nicefrac}       
\usepackage{microtype}      
\usepackage{lipsum}
\usepackage{graphicx}
\graphicspath{ {./images/} }

\usepackage{amsmath}
\usepackage{algorithm}
\usepackage{algorithmic}
\usepackage{amssymb}
\usepackage{mathtools}
\usepackage{amsthm}
\usepackage{amsmath}
\usepackage{mathrsfs}
\usepackage{float}
\usepackage{graphicx}
\newtheorem{theorem}{Theorem}[section]

\newtheorem{lemma}[theorem]{Lemma}
\newtheorem{corollary}[theorem]{Corollary}

\newtheorem{assumption}[theorem]{Assumption}

\usepackage{makecell}
\usepackage{tabularx}
\usepackage{listings}
\usepackage{color}
\usepackage{multirow}
\usepackage{adjustbox}
\usepackage{makecell}
\usepackage{booktabs}
\usepackage{natbib}

\title{Why Federated Optimization Fails to Achieve Perfect Fitting ? \\ A Theoretical Perspective on Client-Side Optima}

\author{
 Zhongxiang Lei \\
  Beijing Institute of Technology\\
  \texttt{zxlei@bit.edu.cn} \\
   \And
 Qi Yang \\
  Beijing Institute of Technology\\
  \texttt{qyang@bit.edu.cn} \\
  \And
 Ping Qiu \\
  Beijing Institute of Technology\\
  \texttt{qiuping@njupt.edu.cn} \\
  \And
 Gang Zhang \\
  Beijing Institute of Technology\\
  \texttt{zhanggang@bit.edu.cn} \\
  \And
 Yuanchi Ma \\
  Beijing Institute of Technology\\
  \texttt{yma@bit.edu.cn} \\
  \And
 Jinyan Liu \\
  Beijing Institute of Technology\\
  \texttt{jyliu@bit.edu.cn} \\
}

\begin{document}
\maketitle
\begin{abstract}
Federated optimization is a constrained form of distributed optimization that enables training a global model without directly sharing client data. Although existing algorithms can guarantee convergence in theory and often achieve stable training in practice, the reasons behind performance degradation under data heterogeneity remain unclear. To address this gap, the main contribution of this paper is to provide a theoretical perspective that explains why such degradation occurs. We introduce the assumption that heterogeneous client data lead to distinct local optima, and show that this assumption implies two key consequences: 
1) the distance among clients’ local optima raises the lower bound of the global objective, making perfect fitting of all client data impossible; and  
2) in the final training stage, the global model oscillates within a region instead of converging to a single optimum, limiting its ability to fully fit the data.
These results provide a principled explanation for performance degradation in non-iid settings, which we further validate through experiments across multiple tasks and neural network architectures. The framework used in this paper is open-sourced at: \url{https://github.com/NPCLEI/fedtorch}.
\end{abstract}

%

\begin{figure*}[!ht]
    \centering
    \includegraphics[width=1\linewidth]{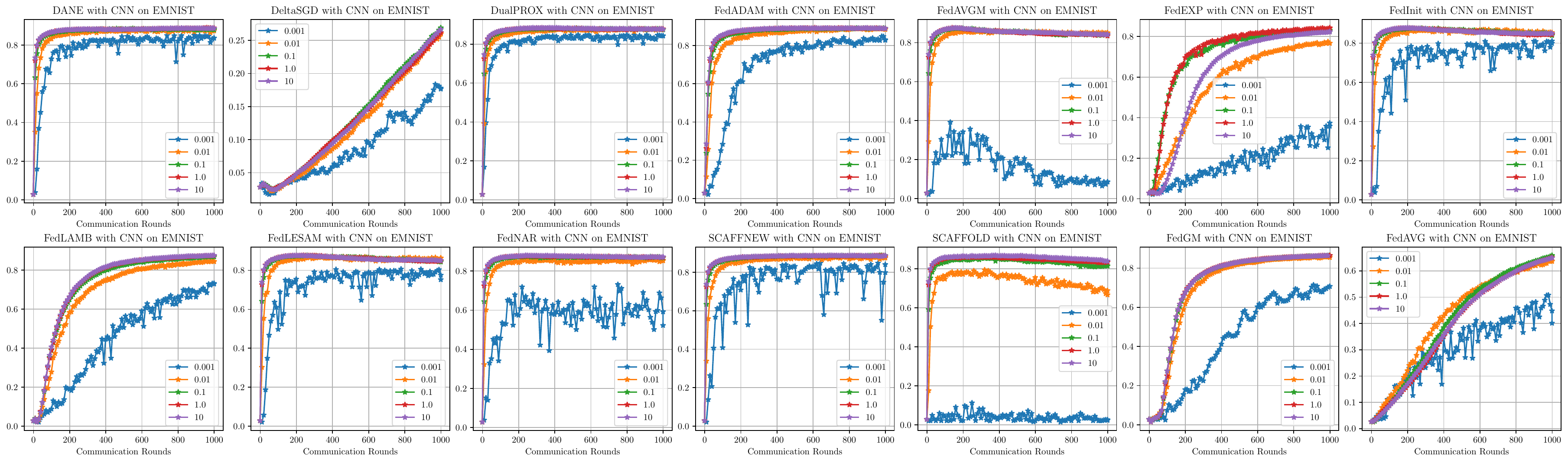}
    \caption{The performance of FedAVG, FedAVGM~\citep{cheng2024momentum}, DeltaSGD~\citep{kim2023adaptive}, FedRed~\citep{jiang2024federated}, FedEXP~\citep{jhunjhunwalafedexp}, FedGM~\citep{sun2024role}, FedInit~\citep{sun2023understanding}, FedLESAM~\citep{qu2022generalized}, FedNAR~\citep{li2023fednar}, FedPROX~\citep{li2020federated}, SCAFFOLD~\citep{karimireddy2020scaffold}, SCAFFNEW~\citep{mishchenko2022proxskip}, and FedADAM~\citep{reddi2020adaptive} on the EMNIST classification task. These experiments demonstrate a common phenomenon: These algorithms converge to the stationary point, but their final performance deteriorates due to increased heterogeneity. Where \(\alpha\) is the Dirichlet distribution parameter, commonly used to simulate heterogeneity.} 
    \label{fig:acc_heter}
\end{figure*}

\section{Introduction}

Federated optimization (FO)~\citep{mcmahan2017communication,reddi2020adaptive} research in a distributed system explores how to optimize machine learning models without requiring clients' data to participate in communications, with neural network models being the most common. The objective function of FO is: 
\begin{equation}
\min_{x \in \mathbb{R}^d} F(x) \quad \text{where} \quad F(x) \stackrel{\text{def}}{=} \frac{1}{|\mathcal{S}|} \sum_{i \in \mathcal{S}} f(x;D_i)~.
\label{eq:target_fl}
\end{equation} 
Here, \(x\) is the machine learning model, \(\mathcal{S}\) is the set of all clients , \( D_i\) is data set of client \(i\) and \(f\) is the loss function of \(x\) on \(D_i\)  (\(f(x;D_i)\) as \(f_i\) for short ). The most classic problem faced by FO is that the distribution of client data \(D_i\) cannot represent the overall distribution, leading to a deterioration in the performance of the optimized model, also known as the non-iid problem.

The main contribution of this paper is to provide a theoretical explanation for the performance degradation of federated optimization under non-iid settings. As illustrated in Fig.~\ref{fig:acc_heter}, although most existing federated optimization algorithms can theoretically guarantee convergence and often achieve stable convergence in practice, the models still exhibit significant underfitting in non-iid scenarios (blue curve), consistently performing worse than in the iid case (purple curve). This phenomenon has not yet been convincingly explained in the current literature. Methods such as SCAFFOLD~\citep{karimireddy2020scaffold} and FedProx~\citep{li2020federated} generally attribute the issue to the “drift” of local client updates; however, this perspective fails to fully capture the root cause of the degradation. Even with the improvements proposed by FedAvgM~\citep{cheng2024momentum}, FedRed~\citep{jiang2024federated}, SCAFFNEW~\citep{mishchenko2022proxskip} and related approaches, experimental results demonstrate that the performance deterioration under non-iid conditions remains persistent.

\subsection{Theoretical conclusions} 

Based on experimental observations, we find that training data across different clients can drive the model to converge to distinct local optima. Motivated by this phenomenon, this paper introduces a novel perspective for \textbf{analyzing the convergence behavior of federated optimization within a single communication round $t$}. This theoretical framework allows us to rigorously establish the following key conclusions:

\begin{itemize}
\item \textbf{Lower Bound from Local Optima Heterogeneity (Theorem~\ref{theorem:ultra})}: Due to heterogeneity among local optima, a significant lower bound exists for Equation~\ref{eq:target_fl}, indicating that the global model cannot perfectly fit the data. This phenomenon may lead to underfitting in non-i.i.d. scenarios. The conclusion is validated experimentally on multiple complex neural networks, including GRU, ResNet-18, ViT, and Deepseek.
\item \textbf{Oscillatory Region (Theorem~\ref{theorem:main})}: During the final stage of convergence, an oscillatory region emerges. Once the global model $x_t$ enters this region, its trajectory exhibits pronounced oscillations, making convergence difficult without gradually decaying the local learning rate to zero. Visual experiments in Fig.~\ref{fig:acc_heter} and several convex function cases corroborate this finding. The theory also elucidates the impact of variable local update rounds, client weights, and participation rates on convergence.
\item \textbf{Trajectory Correction Condition (Theorem~\ref{theorem:LA-FedSGD-Corr})}: Consistent with prior studies~\citep{cheng2024momentum,jiang2024federated}, the effectiveness of the correction term $h_t$ depends on two factors: its norm must not exceed the distance between the global model and the local optimum, and its direction must align with the trajectory.
\item \textbf{Momentum and Adaptive Learning (Theorem~\ref{theorem:SA})}: Our theory offers a fresh perspective on the roles of momentum and learning rate in adaptive methods. Momentum increases the descent distance of a single update, accelerating overall convergence, while adaptive learning rates, though having a decaying effect on distance, can also enhance descent magnitude.
\end{itemize}



\subsection{Research Setting for a Cluster of Federated Optimization Methods}

\begin{algorithm}[ht]
\caption{A cluster of federated optimization methods}
\textbf{Input}: Global step $\eta$ for adaptive optimization to update $x_t$; Local learning rate $\eta_l^t$ to optimize local model; Communication rounds $T$ ; Local update rounds $K_i$. \\
\textbf{Output}: Global model parameter $x$ .

\begin{algorithmic}[1] 
    \STATE Initialization: Global model parameters $x_0$
    \FOR{$t = 0 ,..., T$} 
        \STATE The server connects any number of active clients to form the subset \(\mathcal{S}_t\) and sends \(x_t\) to clients. 
        \FOR{$\textit{i} \in \mathcal{S}_t$ \textbf{in parallel}}
            \STATE  Starting from the point \(x_t\), client \(i\) updates the model \(K_i\) times and obtain the updated state \(x_{i,K}^{t}\) with learning rate \(\eta_l^i\) and optional correction direction \(h_t\).
        \ENDFOR
        \STATE \(x_{t+1} = x_t - \eta_t \sum_{i \in \mathcal{S}_t} \rho_i^t \left( x_t - x_{i,K}^t \right)\).
        \STATE \(\text{s.t.} \sum_{i \in \mathcal{S}_t } \rho_i^t = 1 , \rho_i^t \ge 0\)
    \ENDFOR
    \STATE \textbf{return} $x_T$
\end{algorithmic}
\label{alg:FedNAOMPX}
\end{algorithm}

\noindent
Federated optimization has made significant progress, and the theory proposed in this paper can be shared by abstracting it as much as possible into Algorithm~\ref{alg:FedNAOMPX}. Specifically, based on whether there are correction terms and whether there are adaptive optimization methods, the following three categories of work can be summarized:

\subsubsection{Local adaptive optimization methods (LA)} When the algorithm does not have local correction terms, i.e., \(\eta_t= 1\) and \(h_t = 0\) in Algorithm 1, these algorithms optimize the model locally using some optimization algorithms, such as gradient clipping~\citep{li2024improved} and various adaptive optimization methods~\citep{yangFederatedLearningNesterov2022,mukherjee2023locally}. The most classic method is FedAVG~\citep{mcmahan2017communication}, which uses sgd training on the client side and aggregates samples as weights on the server side.


\subsubsection{Drift correction methods (DC)} Some methods assume that the training trajectories on the client side exhibit “drift,” so they introduce a drift correction term to correct the trajectories. This paper simplifies this drift correction term to \(h_t\). Note that regularization methods can also be viewed as correction terms, such as \(f_i + \| x - x_t \|^2 \), where the correction term can be treated as a correction term when computing the gradient. Representative works include SCAFFNEW, FedAVGM, FedRed, and SCAFFOLD~\citep{mishchenko2022proxskip, cheng2024momentum, jiang2024federated, karimireddy2020scaffold}, which correspond to the case where \(h_t \neq 0\) (with \(\eta\) being arbitrary) in Algorithm 1.



\subsubsection{Server Adaptive optimization methods (SA)} Some studies have introduced highly successful adaptive optimization methods for neural networks into federated learning. For example, FedGM~\citep{sun2024role} uses the Heavy ball and Neserov methods to accelerate training, while FedOPT, FedExp, Fed-EF, and FedAMS ~\citep{reddi2020adaptive,jhunjhunwalafedexp,wang2024fadas,li2023analysis} employ adaptive learning rate methods. They all utilize the pseudo-gradient \(\mathbb{G}_{\mathcal{S}_t} = \sum_{i \in \mathcal{S}_t} \rho_i^t \left( x_t - x_{i,K}^t \right)\) to replace the true gradient in these methods for optimization.

It is worth noting that there are many variations of these methods. This article analyzes the common elements of the above methods and may not cover all versions.

\footnotetext[1]{In these experiments, we optimize the local loss function \(f_i\) to \(1^{-2}\) or the point where the loss does not change anymore as an approximation to the local optimum \(x^*_i\).}

\section{Basic Assumptions of Heterogeneity of Local Optimal Points }
\label{section:basic_assumption}


Our goal is to propose a theoretical analysis that focuses solely on the parameters of federated learning, such as heterogeneity, local training rounds, client participation rate, and aggregation weights, etc. The challenge is how to formulate an assumption that covers all settings for analysis, given that the data distribution on clients and the function properties are unknown. A key observation is that: \textbf{the client's data \(D_i\) differs from other clients}, which implies that for clients \(i\) and \(j\), the parameters that minimize their respective target functions \(f_i\) and \(f_j\) are distinct; in other words, most their local optimal points satisfy \(x_i^* \ne x_j^*\). We conduct extensive experiments to verify this conjecture. In Fig. \ref{fig:assumption1_re}(left), we plotted the loss landscape on the client side and the relative positions of the approximated optimal points of the neural network model at different rounds (Fig. \ref{fig:assumption1_re}(right)). The experiments demonstrated that there are multiple local optima on the client side. 
Based on this, we formally propose the following assumption:

\begin{figure*}[ht]
    \centering
    \includegraphics[width=1\linewidth]{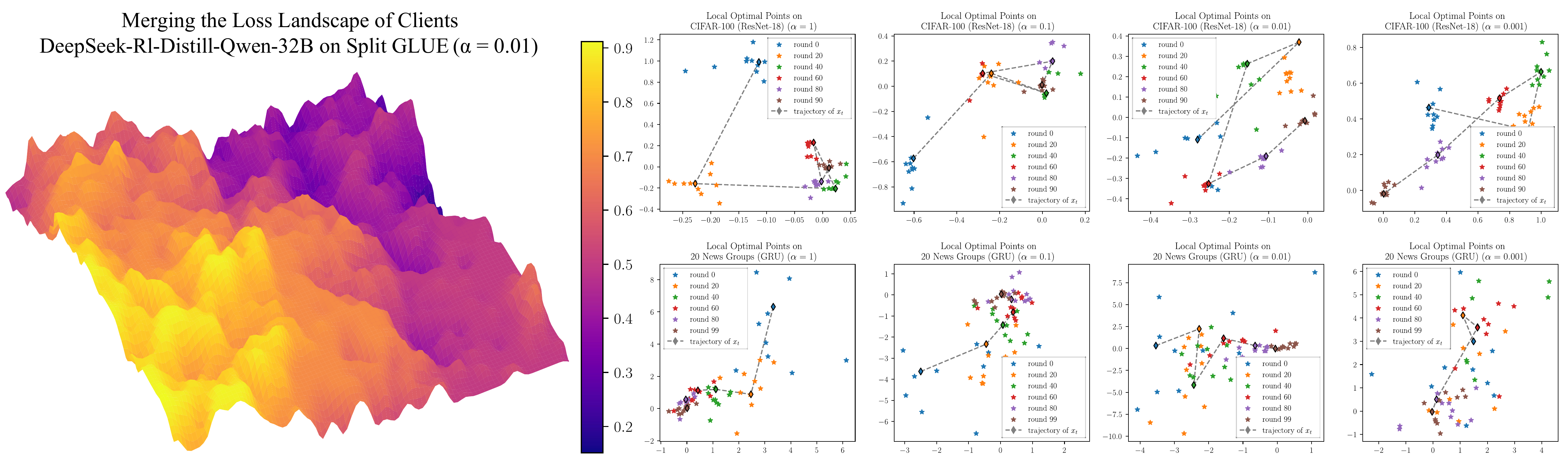}
    \caption{The two figures on the left and right support the view that the local optimal points of the clients is heterogeneous. The left figure merges the loss landscapes of all \(f_i\) into a single plot using the aggregation method \(g(x) = \min (f_1(x), f_2(x), f_{\dots}(x))\). The right figure demonstrates the changes in the relative positions of local optimal points (approximation\protect\footnotemark[1]) under different tasks and rounds.}
    \label{fig:assumption1_re}
\end{figure*}


\begin{assumption} (Heterogeneity of Local Optimal Points)
\label{assumption:local_optimal_heter}
Let that \( x_{i}^* = \arg_x \min f(x;x_t,\mathcal{D}_i) \) is the local optimal point of \( f_i \) achieved by client \( i \) start optimization at \( x_t \). The average distance of local optimal points satisfies the following relationship:
\begin{equation}
\mathcal{H}(\{x^*\}_{i \in \mathcal{S}}) = \frac{1}{\left| \mathcal{S} \right|} \sum_{i \in \mathcal{S}}  \|x_{i}^* - x_{\mathcal{S}}^* \|^2 \ge 0 ,
\label{eq:assumption of heter}
\end{equation} where \(x_{\mathcal{S}}^* = \frac{1}{\left| \mathcal{S} \right|} \sum_{i \in \mathcal{S}} x_i^* \) is center of local optimal points. The equality holds if and only if all client data are i.i.d. It should be noted that in non-convex settings, $x_i^*$ may change as $x_t$ changes. Unlike convergence analysis, the conclusions in this paper focus only on the behavior at time $t$, so the change of $x_i^*$ does not affect the subsequent results.



\begin{assumption}(Locally Efficient Descent Condition) \label{assumption:eff_dec} Define the initial vector pointing to the local optimum as \( \delta_{t,i} = x^{t} - x_{i}^* \). Let \( x_{i,K}^{t} \) represent the updated model after client-side updates. The updated vector pointing to the local optimum is then defined as \( \delta_{i,K}^{t} = x_{i,K}^{t} - x_{i}^* \). A client \( i \) is considered effectively descending if it satisfies the condition:
\[
\left\| \delta_{t,i}^K \right\| = \sigma_{i}^{t}(K,\eta_l) \left\| \delta_{t,i} \right\|,  0 \leq \sigma_{i}^{t}(K,\eta_l) \leq 1,
\] where \(\sigma_i^t(K,\eta_l)\) (\(\sigma_i^t\) for short) comprehensively incorporates properties of \(f_i\) and some properties related to the optimization method used by client \(i\).

Assumption \ref{assumption:eff_dec} effectively bypasses assumptions about the properties of function \(f_i\) and its gradient in \(\sigma_i^t(K,\eta_l)\), allowing us to focus more on the parameters of FL rather than the specific properties of the function.
For example, \( \sigma_{i}^t \) usually has a relationship where the larger \( K_i \) is, the closer \( \sigma_{i}^t \) is to 0. 
For methods using adaptive optimization neural networks, \(\sigma_{i}^t\) is related to the descent accuracy.
For methods that use adaptive optimization methods to optimize neural networks, \(\sigma_{i}^t\) is related to the descent precision. For example, when using adaptive methods such as Adam~\citep{defossez2020simple,kingma2014adam} and Adagrad~\citep{li2023convergence}, \(\sigma_{i}^t = \mathcal{O}( K^{-1/2} )\) is below this order of magnitude.
And ordinary gradient methods can guarantee that \( \left\| \delta_{t,i}^K \right\| \leq ( \frac{Q_f - 1}{Q_f + 1} )^K \left\| \delta_{t,i} \right\| \) where \( Q_f \) is related to the properties of the function (strong convexity \( \mu \), smoothness \( L\), etc.). Nesterov's acceleration method~\citep{nesterov1983method,assran2020convergence} can guarantee that \( \sigma_{i}^t = \mathcal{O}( K^{-2} ) \). For general smooth cases, often \( \sigma_{i}^t = \mathcal{O}( K^{-1/2} ) \), which can all be found in Nesterov's Lectures ~\citep{nesterov2018}. 
\end{assumption}
\end{assumption}

\section{Lower Bound from Local Optima Heterogeneity}

\begin{figure}[h!]
    \centering
    \includegraphics[width=0.85\linewidth]{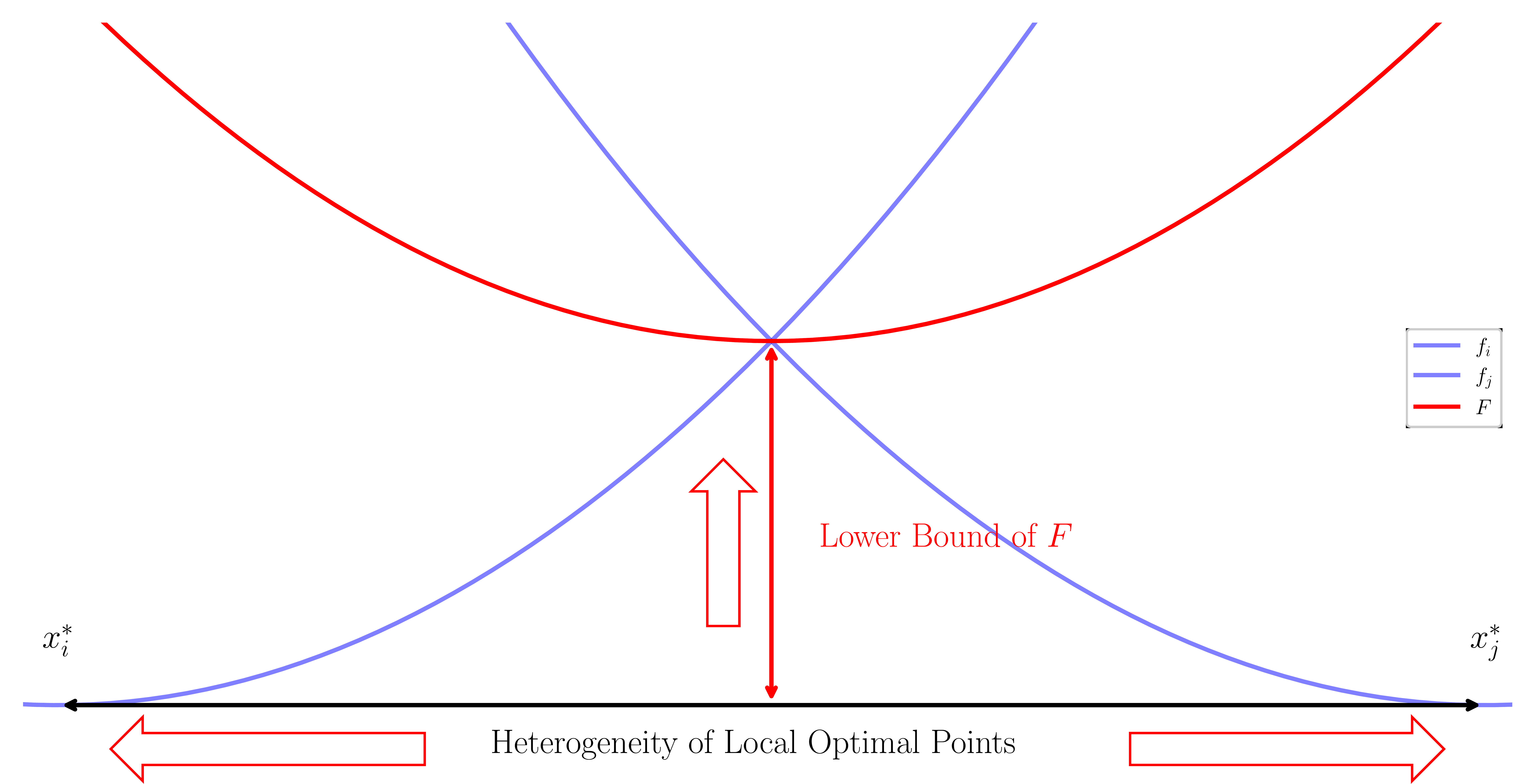}
    \caption{The lower bound of Eq.~\ref{eq:target_fl}(F) will be pulled up by distant of local optimal points.}
    \label{fig:enter-label}
\end{figure}

Although neural networks are inherently non-convex, their loss landscape near optimal points can be approximated as convex. Theoretical support comes from two sources: 1) Under certain conditions \citep{286886,118640}, neural networks are universal function approximators, enabling clients to reach local optima \(f(x_i^*)\); and 2) Visualization studies \citep{li2018visualizing} reveal that near these optima, ResNets exhibit convex-like loss landscapes characterized by positive Hessian eigenvalues.

\begin{theorem}
\label{theorem:ultra}(Lower Bound of Objective Function) If \( f(x; D_i) \) can be approximated as a convex function within the neighborhood \( U_i(x_i^*) \) around \( x_i^* \), then \( \nabla f_i = 0 \) and \( \nabla f_i^* \) is positive definite. For all \( x \in U_0 \cap U_1 \cap \ldots \), the lower bound of Eq. \ref{eq:target_fl} is:
\begin{equation}
F(x) \ge \frac{1}{|\mathcal{S}|} \sum_{i \in \mathcal{S}} \left(  f(x_i^*) + \frac{\lambda_{\text{min}}^i }{2} \Bigg( \| x_i^* - x_{\mathcal{S}}^* \| - \left\|x - x_{\mathcal{S}}^* \right\| \Bigg)^2 \right)    
\end{equation} where \(\lambda_{\text{min}}^i > 0\) is the smallest eigenvalue of \(\nabla^2 f_i(x_i^* + t (x - x_i^*)) , t \in (0,1)\). 

\begin{proof}
Apply the function by Taylor's theory \citep{boyd2018introduction} to \(f_i\):

\begin{align*}
&f_i(x_i^* + (x - x_i^*)) = f_i(x_i^*) + (x - x_i^*)^\top \nabla f_i(x_i^*)  +  \frac{1}{2} (x - x_i^*)^\top \nabla^2 f_i(z_i) (x - x_i^*) \\
&= f_i(x_i^*) + \frac{1}{2} (x - x_i^*)^\top \nabla^2 f_i(z_i) (x - x_i^*) \\
&= f_i(x_i^*) + \frac{1}{2} \left((x - x_{\mathcal{S}}^*) - ( x_i^* - x_{\mathcal{S}}^* ) \right)^\top \nabla^2 f_i(z_i) \left((x - x_{\mathcal{S}}^*)   - ( x_i^* - x_{\mathcal{S}}^* ) \right) \\
&\ge f_i(x_i^*) + \frac{\lambda_{min}^i}{2} \left\|(x - x_{\mathcal{S}}^*) - ( x_i^* - x_{\mathcal{S}}^* ) \right\|^2 \\
&\ge f_i(x_i^*) + \frac{\lambda_{min}^i}{2} \Bigg| \left\|(x - x_{\mathcal{S}}^*)\right\| - \left\| ( x_i^* - x_{\mathcal{S}}^* ) \right\| \Bigg|^2
\end{align*} where \(\lambda_{\text{min}}^i > 0\) is the smallest eigenvalue of the positive definite matrix \(\nabla^2 f_i(x_i^* + t (x - x_i^*)) , t \in (0,1)\).
\end{proof} 
\end{theorem}

\subsection{Results Analysis and Experimental Validation}


\begin{table*}[th!]
\small
\setlength{\tabcolsep}{1.23mm}
\centering
\begin{tabular}{rllllllllllll}
\hline
\multicolumn{1}{c}{\multirow{2}{*}{\(\alpha\)}} & \multicolumn{3}{l}{\begin{tabular}[c]{@{}l@{}}GRU on 20 \\ NewsGroups\end{tabular}} & \multicolumn{3}{l}{\begin{tabular}[c]{@{}l@{}}Resnet 18 on \\ CIFAR100\end{tabular}} & \multicolumn{3}{l}{\begin{tabular}[c]{@{}l@{}}Vit-large on\\ ImageNet 1K\end{tabular}} & \multicolumn{3}{l}{\begin{tabular}[c]{@{}l@{}}DeepSeek-R1-32B\\ on GLUE\end{tabular}} \\
\multicolumn{1}{c}{}                            & \(F^*\)                 & acc.                     & \(\mathcal{H}\)                & \(F^*\)                 & acc.                     & \(\mathcal{H}\)                 & \(F^*\)                  & acc.                      & \(\mathcal{H}\)                 & \(F^*\)                  & acc.                     & \(\mathcal{H}\)                 \\ \hline
0.001                                           & 0.79                    & 44.19 \%                 & 0.63                           & 1.18                    & 47.86 \%                 & 1.93                            & 0.59                     & 77.70 \%                  & 0.24                            & 1.62                     & 60.45 \%                 & 1.57                            \\
0.01                                            & 0.35                    & 64.67 \%                 & 0.39                           & 0.54                    & 52.57 \%                 & 1.08                            & 0.44                     & 79.87 \%                  & 0.19                            & 1.73                     & 70.86 \%                 & 1.16                            \\
0.1                                             & 0.04                    & 74.70 \%                 & 0.21                           & 0.50                    & 54.16 \%                 & 0.97                            & 0.33                     & 80.34 \%                  & 0.17                            & 0.79                     & 78.10 \%                 & 0.82                            \\
1                                               & 0.03                    & 75.16 \%                 & 0.20                           & 0.22                    & 57.41 \%                 & 1.04                            & 0.33                     & 80.71 \%                  & 0.14                            & 0.48                     & 81.24 \%                 & 0.73                            \\
10                                              & 0.03                    & 74.83 \%                 & 0.17                           & 0.21                    & 56.97 \%                 & 1.10                            & 0.33                     & 80.72 \%                  & 0.15                            & 0.42                     & 82.86 \%                 & 0.71                            \\ \hline
\(Per(*,\mathcal{H})\)                          & 0.997                   & -0.990                   & -                              & 0.895                   & -0.838                   & -                               & 0.962                    & -0.971                    & -                               & 0.888                    & -0.997                   & -                               \\ \hline
\end{tabular}
\caption{Approximate value of \(F(x^*)\) (\(F^*\)), test accuracy (acc.), \(\mathcal{H}(\{x^*\}_{i \in \mathcal{S}}) (\mathcal{H}) \) and the parameter \(\alpha\) of the Dirichlet distribution  exhibit correlations across various neural network tasks.\(Per(*,\mathcal{H})\) represents the Pearson correlation coefficient between the test accuracy column and the \(\mathcal{H}\) column, or between the \(F^*\) column and the \(\mathcal{H}\) column.}
\label{tab:afh}
\end{table*}

The lower bound of the objective function Eq.~\ref{eq:target_fl} of FL is bounded by two inherent limitations: the first is the degree to which the local function \( f(x_i^*) \) is fitted, and the second is the heterogeneity of the local optimal points, as defined in Assumption \ref{assumption:local_optimal_heter}. Even if a client fully fits its local data set such that \( f(x_i^*; D_i) = 0 \), the lower bound of the objective function will still be constrained by the distance between the local optima, preventing Eq \ref{eq:target_fl} from perfectly fitting all data \( \{D\}_{i \in \mathcal{S}} \). As long as Assumption \ref{assumption:local_optimal_heter} is not equal to zero, Eq. \ref{eq:target_fl} is unlikely to be zero. 

To verify whether this theory is effective for neural networks, we designed an experiment that was conducted on a variety of complex tasks and neural network structures (including shallow RNNs, Resnet-18, large Vision Transformers ~\citep{radford2021learning} , and the recently popular large network Deepseek-R1-32B~\citep{guo2025deepseek}). 
The experimental results presented in Table~\ref{tab:afh} demonstrate a correlation among the objective function value \(F(x^*)\), the distant of local optimal points \(\mathcal{H}(\{x^*\}_{i \in \mathcal{S}})\), test accuracy (acc.), and \(\alpha\), thereby providing empirical validation for Theorem \ref{theorem:ultra}. This lower bound also explains why FL can sometimes prevent overfitting, but at other times it can lead to underfitting.

\section{Heterogeneity Theorems of Local Optimal Points}


This section is divided into three theories, which are presented in a step-by-step manner. Theorem~\ref{theorem:main} analyzes the single-step convergence conclusions when arbitrary optimization methods are used on the client side and the FedAVG algorithm is used on the server side, and provides separate analyses of trajectory analysis, the role of the local update times K, the role of weights, and the role of participation rates. Theorem~\ref{theorem:LA-FedSGD-Corr} examines the effect of the correction term \(h_t\) on single-step convergence when it is present on the client side. Theorem~\ref{theorem:SA} analyzes the roles of momentum and adaptive learning rate in SA for single-step updates.

\subsection{Theorem of LA-FedAVG: Trajectory and Role of Factors Analysis}


The advantage of Assumption~\ref{assumption:eff_dec} is that it allows analysis when participating clients optimize models using different optimization methods, because different optimization methods can ultimately calculate different \(\sigma_{i}^{t}(K,\eta_l)\). Then, when clients optimize using any optimization method and there is no correction term (i.e., \(h_t\)), the Local adaptive optimization methods (LA) mentioned in Chapter 2 have the following theoretical analysis:

\begin{theorem} 
\label{theorem:main}
(LA-FedAVG) Assuming that local optimal points satisfy Assumption \ref{assumption:local_optimal_heter} and all clients \( i \in \mathcal{S}_t \) satisfy the effective descent condition~\ref{assumption:eff_dec}. The single update distance from \( x_{t+1} \) of LA-FedAVG  to the weighted-sampled local optimal point \( x^*_{\mathcal{S}_t} = \sum_{i \in \mathcal{S}_t} \rho_i x_i^* \) is:
\begin{equation}
\label{eq:main}
\begin{split}
\left\| x_{t+1} - x^*_{\mathcal{S}_t} \right\|^2  = \ &\left\| x_{t} - x^*_{\mathcal{S}_t} \right\|^2  - \frac{1}{|\mathcal{S}_t|^2} \left(\mathcal{P}_{\mathcal{S}_t} \cdot X_{\mathcal{S}_t} \right)^\top A  \left(\mathcal{P}_{\mathcal{S}_t} \cdot X_{\mathcal{S}_t} \right),
\end{split}
\end{equation}
where \(\mathcal{P}_{\mathcal{S}_t} = \left[\rho_i^t\right]_{i \in \mathcal{S}_t} \) is weight vector, \(X_{\mathcal{S}_t} = \left[ \delta_{t,i} \right]_{i \in \mathcal{S}_t} \), and the element of matrix A  is \( A_{i,j} = \cos \left\langle  \delta_{t,i},  \delta_{t,j} \right\rangle - \sigma_i^t \sigma_j^t  \cos \left\langle  \delta_{t,i}^K, \delta_{t,j}^K \right\rangle \). 
\end{theorem}



\subsubsection{Trajectory Analysis}

\begin{figure*}[ht]
    \centering
    \includegraphics[width=1\linewidth]{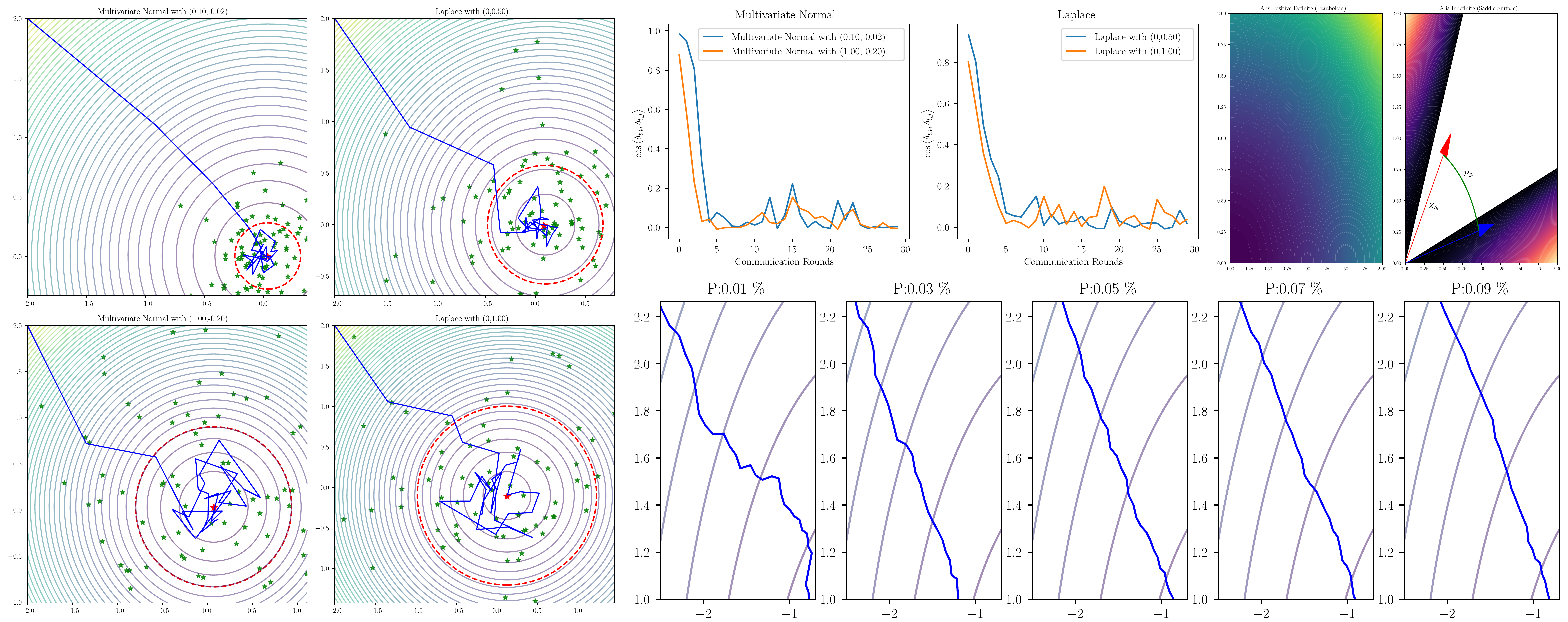}
    \caption{The left four diagrams represent the contour plots of paraboloid surface \( f \), where the green points indicate the local optimal point of \( f_i \) obtained by sampling  using both multivariate Gaussian and Laplace distributions. The blue line shows the optimization trajectory of \( x_t \), while the red dashed lines outline the oscillatory region. In the upper right corner, the left plot shows the communication rounds on the horizontal axis and the average value of \(\cos \left\langle  \delta_{t,i},  \delta_{t,j} \right\rangle\) on the vertical axis. The two contour plots in the upper right corner explain the effect of the positive definiteness of \( A \) on the range of quadratic forms. The white areas indicate regions where the values are less than zero, the red vectors represent \( X_{\mathcal{S}_t} \) before being deflected by \( \mathcal{P}_{\mathcal{S}_t} \), and the blue vectors represent the deflected ones. The lower right section presents the trajectories for \( P \) ranging from 0.01\% to 0.09\%.}
    \label{fig:main_theorem}
\end{figure*}

To analyze whether \( x_{t+1} \) moves closer to \( x^*_{\mathcal{S}_t} \) than \( x_t \), the key is to analyze when the quadratic form is positive definite: \( \Delta_{t+1} (X_{\mathcal{S}_t}) = \frac{1}{|\mathcal{S}_t|^2} \left(\mathcal{P}_{\mathcal{S}_t} X_{\mathcal{S}_t} \right)^\top A  \left(\mathcal{P}_{\mathcal{S}_t} X_{\mathcal{S}_t} \right) > 0 \). Analyzing this matrix as a whole is extremely complex, particularly due to the interdependence of the vector angles and the fact that research ~\citep{vershynin2009high} considers the expectation of quadratic forms to be chaotic, making it difficult to establish precise conditions for \(A\) being positive definite. However, we can still observe the characteristics of matrix elements to analyze trajectory information.

We observe that \(\mathcal{P}_{\mathcal{S}_t} X_{\mathcal{S}_t} \ge 0 , A_{i,i} = 1 - \sigma_{i,t}^2 \ge 0\) and \(A\) is the Hessian matrix of quadratic form.  Hence \(A\) directly determines whether the surface of the quadratic function is convex or non-convex (in the top right corner of Fig.\ref{fig:main_theorem}).  In particular, if the quadratic form is everywhere convex, then \(A\) is positive semi-definite; If the quadratic form exhibits non-convexity, it may nevertheless remain nonnegative in regions close to the coordinate axes :
\(
\lim_{\substack{X_{\mathcal{S}_t}[i]\to 0\\i\neq j}}\Delta_{t+1}
\;=\;A_{j,j}\,(\mathcal{P}_{\mathcal{S}_t}[j]\cdot X_{\mathcal{S}_t}[j])^2\;\ge\;0
\). From a numerical standpoint, whenever the following inequality holds:
\(
\sum_{i\in\mathcal{S}_t}\rho_i(1-\sigma_{i,t}^2)\|\delta_{t,i}\|^2
\;\ge\;
\sum_{i\in\mathcal{S}_t}\sum_{\substack{j\in\mathcal{S}_t\\j\neq i}}
\rho_i\,\rho_j\!(\sigma_{i,t}\,\sigma_{j,t}\,\cos\langle\delta_{t,i}^K,\delta_{t,j}^K\rangle
\;-\;\cos\langle\delta_{t,i},\delta_{t,j}\rangle)\,\|\delta_{t,i}\|\,\|\delta_{t,j}\|,
\)
The quadratic form is positive semi-definite. Because of \(A_{i,i} \ge 0\), we can only conclude that \(A\) is likely positive definite when "most" of its off-diagonal elements satisfy \( A_{i,j} \ge 0 \Rightarrow \cos \left\langle  \delta_{t,i},  \delta_{t,j} \right\rangle  \ge \sigma_i^t \sigma_j^t  \cos \left\langle  \delta_{t,i}^K, \delta_{t,j}^K \right\rangle \). 
The above information is enough for us to analyze the optimization trajectory of \(x_t\). Ignoring \(\sigma_i^t\) and \(\sigma_j^t\) temporarily, if the local optimal points do not change with variations in \(x_t\), then we can deduce that:

\textit{Stage 1}: When \(x_t\) is far away from both \(x_i^*\) and \(x_j^*\), we can obtain  \(\cos\left\langle \delta_{t,i}, \delta_{t,j} \right\rangle \approx 1 => A_{i,j} \ge 0\). In this case, \(A\) is semi-positive definite and \(x_{t+1}\) approaches \(x^*_{\mathcal{S}_t}\). 

\textit{Stage 2}: As \(x_t\) gradually moves closer to \(x_i^*\) and \(x_j^*\), with the angle between them increasing, \(A_{i,j} > 0\) no longer holds for most elements, making matrix \(A\) indefinite. Consequently, \(x_{t+1}\) begins to move away from \(x^*_{\mathcal{S}_t}\). 

\textit{Stage 3}: Once \(x_t\) moves far away again from \(x_i\) and \(x_j\), the angle between them decreases, and \(x_{t+1}\) starts to approach \(x^*_{\mathcal{S}_t}\) once more. This indicates that \(x_t\)'s trajectory begins to \textbf{oscillate} around \(x^*_{\mathcal{S}_t}\). 

\subsubsection{Oscillatory region  and Heterogeneity}
When the mean vector angle between \(x_t\) and \(x_i^*\), \(x_j^*\) is greater than or equal to 90 degrees, it forms a spherical region with \(\|x_{t,i}^* - x_{t,j}^*\|\) as its diameter, we call this region is \textbf{oscillatory region}:
\begin{equation}
\left\| x_t - x_{\mathcal{S}}^* \right\|  = \frac{1}{2}\left\| x_j^* - x_i^* \right\| \text{~if~}  \cos \left\langle  \delta_{t,i},  \delta_{t,j} \right\rangle =0. \nonumber
\end{equation}
As shown by the red dashed line (centered at \( x_{\mathcal{S}}^* \) with radius \( \frac{1}{2|\mathcal{S}| (|\mathcal{S}| - 1)} \sum_{i \in \mathcal{S}} \sum_{\substack{j \in \mathcal{S} \\ i \ne j}} \left\| x_j^* - x_i^* \right\| \)) on the left side of Fig. \ref{fig:main_theorem}, the experimental results on the paraboloid fully support our analysis of the trajectories. Once \(x_t\) enters this region of the sphere, \(\sigma_i^t\) and \(\sigma_j^t\) become small enough to fail (refer to The Role of \(K_i\)).
At this point, even if the client updates faster, it becomes ineffective, causing the trajectory of \(x_t\) to begin oscillating. 
In the worst case, this can lead to \(A\) becoming indefinite, resulting in the trajectory diverging outward. 
This also explains why gradient heterogeneity theories suggest that the local learning rate \(\eta_l = \mathcal{O}(\frac{1}{T})\) is required.

\subsubsection{The Role of Loacl Update Rounds \texorpdfstring{$K_i(\sigma_{i}^t)$}{Ki(sigma\_i(t))}} 
One of the original intentions behind FL ~\citep{mcmahan2017communication} design is to increase the local training times \( K_i \) of clients to save communication times and costs. If the local optimum point \(x_{i,t}^*\) is fixed (e.g. \(f_i\) is strong convex) and 
unchanged, the smaller \(\sigma_i^K\) (the larger \(K\)) before \(x_t\) enters the oscillatory region, the better. However, in the case of neural networks, the local optimum point will change with the change of \(x_t\), and a larger \(K\) may make the algorithm prematurely fall into the poor-performance optimal point.

\subsubsection{The Role of Weights} There are two places where weights influence Eq. \ref{eq:main}. The first is in the calculation of \( x^*_{\mathcal{S}_t} \), which computes the weight center of the local optimal points in \( \mathcal{S}_t \) and affects the descent direction of \( x_{t+1} \). The second is \( \mathcal{P}_{\mathcal{S}_t} \cdot X_{\mathcal{S}_t} \), where since the sum of the weight vector is 1, its role is to deflect the angle of the vector \( X_{\mathcal{S}_t} \) by a certain degree Fig. \ref{fig:main_theorem} (top right corner). This determines whether the descent distance of \( x_{t+1} \) is more biased towards which client :\(\lim_{\substack{\mathcal{P}_{\mathcal{S}_t}[i] \to 1 \\ i \ne j }} \Delta_{t+1} = A_{j,j} \left(  X_{\mathcal{S}_t}[j]\right)^2 \ge 0 \).

\subsubsection{The Role of Participation Rate} In Theorem \ref{theorem:main}, the main term that participates in this participation rate is \( x_{\mathcal{S}_t}^* \). This term represents an estimate of the weighted center \( x_{\mathcal{S}}^* \) of all local optima. As shown in the lower right corner of Fig. \ref{fig:main_theorem}, if the participation rate is too low, it can lead to a certain bias in each estimation, which may result in oscillations in the trajectory. A high participation rate, on the other hand, does not affect the estimate of the weighted center.  


\subsection{Theorem of Drift Correction Methods }


When the client side uses the SGD method to optimize the model and there is a correction term \(h_t \neq 0\), the DC method in Chapter 2 has the following theoretical analysis when the server side uses SGD to update the global model.

\begin{lemma} \label{lemma:pseudo-gradient} Pseudo-gradient \(\mathbb{G}_{\mathcal{S}_t} = \sum_{i \in \mathcal{S}_t} \rho_i^t \left( x_t - x_{i,K}^t \right) 
 \) can be regarded as the update direction of the LA-FedAVG algorithm: \(
x_{t+1}^{\text{LA-FedAVG}} = x_t - \sum_{i \in \mathcal{S}_t} \rho_i^t \left( x_t - x_{i,K}^t \right)  \Rightarrow x_{t+1}^{\text{LA-FedAVG}} - x_t =  -  \mathbb{G}_{\mathcal{S}_t} \). If the matrix \(A\) is positive definite, then the pseudo-gradient direction points in the direction that takes \(x_{t+1}\) away from \(x_{\mathcal{S}_t}\).
\end{lemma}

\begin{theorem} (DC-FedSGD) 
\label{theorem:LA-FedSGD-Corr}
Assuming the clients satisfy Assumptions \ref{assumption:local_optimal_heter} and \ref{assumption:eff_dec}, and the updating model takes the form of \(x_{i,K}^{h,t} = x_t - \eta_l^i \sum_{i=0}^{K_i-1} \nabla f_i + \eta_l^i K_i h_t = x_{i,K}^t + h_t^i\). Let \(\delta_{\mathcal{S}_t}^K = x_{t+1} - x^*_{\mathcal{S}_t}\) and \(\delta_{\mathcal{S}_t} = x_t - x^*_{\mathcal{S}_t}\), the single-round update distance is:
\begin{equation}
\begin{split}
\left\| x_{t+1} - x^*_{\mathcal{S}_t} \right\|^2  = \
&(1 - \eta_t\wp) \left\| x_{t} - x^*_{\mathcal{S}_t} \right\|^2  - \eta_t\left( \eta_t\Delta_{t+1} + \eth \left\| \mathbb{H}_{\mathcal{S}_t}   \right\| \right),
\end{split}
\end{equation}  where \( \wp = 2(1 - \eta_t) \left( 1 - \cos \langle \delta_{\mathcal{S}_t}^K , \delta_{\mathcal{S}_t} \rangle \sigma_\Delta  \right) , \eth = [  2 \hbar \left\| \delta_{\mathcal{S}_t}  \right\| - \eta_t \left\| \mathbb{H}_{\mathcal{S}_t}  \right\| ] , \hbar =  - \eta_t\sigma_{\Delta} \cos \langle \delta_{\mathcal{S}_t}^K , \mathbb{H}_{\mathcal{S}_t}  \rangle + (\eta_t- 1) \cos \langle \delta_{\mathcal{S}_t} , \mathbb{H}_{\mathcal{S}_t}  \rangle  \) and \(\sigma_{\Delta_{t+1}} = \sqrt{\left\| \delta_{\mathcal{S}_t}  \right\| - \Delta_{t+1}}/\left\| \delta_{\mathcal{S}_t}  \right\|  , \mathbb{H}_{\mathcal{S}_t} = \sum_{i \in \mathcal{S}_t} \rho_i h_t^i \).

\begin{corollary}
\label{corollary:correct-term} (Effective Condition of \( \mathbb{H}_{\mathcal{S}_t} \))
Whether the correction direction \( \mathbb{H}_{\mathcal{S}_t} \) is absolutely effective depends on if \( \eth > 0 \Rightarrow \left\| \mathbb{H}_{\mathcal{S}_t} \right\| < 2 \frac{\hbar}{\eta_t} \left\| \delta_{\mathcal{S}_t}  \right\| \text{~and~}  \frac{1 - \eta_t}{\eta_t\sigma_\Delta} \cos \langle \delta_{\mathcal{S}_t} , \mathbb{H}_{\mathcal{S}_t} \rangle <  \cos \langle \delta_{\mathcal{S}_t}^K , \mathbb{H}_{\mathcal{S}_t}  \rangle \). This conclusion intuitively shows that the norm of \( H_{S_t} \) should not exceed \( \left\| \delta_{\mathcal{S}_t} \right\| \), and it is optimal when its direction is opposite to \( \delta_{\mathcal{S}_t} \). 
\textbf{As far as we know, this theorem is the first to provide the effective condition of \( \mathbb{H}_{\mathcal{S}_t} \), and this result is suitable for papers ~\citep{cheng2024momentum,mishchenko2022proxskip,karimireddy2020scaffold} with a correct term.}
\end{corollary}

\textit{Use `Implicit Correct-Term' to expand the analysis}: Theorem~\ref{theorem:LA-FedSGD-Corr} has strong extensibility and can be used to analyze other methods and properties by replacing correct-direction \(h_t\). For example, training neural networks typically employs a batch manner, where sampling from the data set inevitably leads to inaccurate gradient estimation. For instance, let $\xi_B \sim D_i$, then we have:\(\| \nabla f(\mathbf{x}_{t,i}^k; \xi_B) - \nabla f (\mathbf{x}_{t,i}^k; D_i) \|^2 = \varsigma_{t,i}^k,\) where \(\varsigma_{t,i}^k\) represents the noise introduced due to sampling imprecision. If we regard this noise as \(h_i^t\) (although its role differs from ``correction direction''), the analysis under Theorem \ref{theorem:LA-FedSGD-Corr} still holds, and Corollary \ref{corollary:correct-term} remains applicable.

\begin{proof}
\textbf{Proof idea :} Although we have summarized the commonly used federated optimization methods in Algorithm \ref{alg:FedNAOMPX}, it is impossible for us to analyze all methods in one article. We will use a case study to demonstrate how the conclusions of Theorem \ref{theorem:main} can be used to simply analyze methods not discussed in this paper and future theoretical approaches.

Through Lemma \ref{lemma:pseudo-gradient}, we can know that the pseudo-gradient and the LA-FedAVG algorithm vector are consistent. For the starting operation of LA-FedSGD-Corr (referred to as LAFC for short), it involves substituting the pseudo-gradient using \( x_{t+1}^{\text{LA-FedAVG}} - x_t = -\mathbb{G}_{\mathcal{S}_t} \) to establish a connection with Theorem \ref{theorem:main}.
Assume client update as follows:
\(
x_{i,K}^{h,t} = x_{t} - \eta_l^i \sum_{i=0}^{K^i-1} \nabla f_i + \eta_l^i K^i h_t = x_{i,K}^{t} +  h_t^i 
\). The update \(x_{t+1}^{\text{LAFC}}\) of Algorithm~\ref{alg:FedNAOMPX} is \(
x_{t+1}^{\text{LAFC}} - x_t  = - \eta_t\left[ \sum_{i \in \mathcal{S}_t} \rho_i (x_t - x_{i,K}^{t} -  h_t^i) \right]  = - \eta_t\left(\mathbb{G}_{\mathcal{S}_t} - \mathbb{H}_{\mathcal{S}_t} \right)
\) where \(  \mathbb{G}_{\mathcal{S}_t} = \sum_{i \in \mathcal{S}_t} \rho_i (x_t - x_{i,K}^{t}) \) and \( \mathbb{H}_{\mathcal{S}_t} = \sum_{i \in \mathcal{S}_t} \rho_i h_t^i \).
\begin{equation}
x_{t+1}^{\text{LAFC}} - x_{\mathcal{S}_t}^* = \eta_t\left( x_{t+1}^{\text{LA-FedAVG}} - x_{\mathcal{S}_t}^* \right) + (1 -\eta_t) ( x_t - x_{\mathcal{S}_t}^*) + \eta_t\mathbb{H}_{\mathcal{S}_t} \nonumber
\end{equation} For the equation above, taking the 2-norm on both sides:
\begin{align*}
\left\| x_{t+1}^{\text{LAFC}} - x_{\mathcal{S}_t}^* \right\|^2 \
&= \left\| \eta_t\left( x_{t+1}^{\text{LA-FedAVG}} - x_{\mathcal{S}_t}^* \right) + (1 -\eta_t) ( x_t - x_{\mathcal{S}_t}^*) + \eta_t\mathbb{H}_{\mathcal{S}_t} \right\|^2\\
&= \left( 2 \eta^2_t - 2 \eta_t+ 1 + 2 \eta_t(1 - \eta_t) \cos \langle \delta_{\mathcal{S}_t}^K , \delta_{\mathcal{S}_t} \rangle \sigma_\Delta  \right) \left\| \delta_{\mathcal{S}_t}  \right\|^2  - \eta^2_t \Delta + \eta^2_t \left\| \mathbb{H}_{\mathcal{S}_t}  \right\|^2 \\
&+ \left[ 2 \eta^2_t \sigma_\Delta \cos \langle \delta_{\mathcal{S}_t}^K , \mathbb{H}_{\mathcal{S}_t}  \rangle + \right. \\
& \left. \eta_t(1 - \eta_t) \cos \langle \delta_{\mathcal{S}_t} , \mathbb{H}_{\mathcal{S}_t}  \rangle \right] \left\| \delta_{\mathcal{S}_t}  \right\| \left\| \mathbb{H}_{\mathcal{S}_t} \right\|. 
\end{align*} By organizing the equations, we can prove the conclusion. 
\end{proof}
\end{theorem}

\subsection{Theorem of SA}

For SA with any local optimization method, we refer to the form of QHM ~\citep{gitman2019understanding}:
\(d^t = (1 - \beta_t) \mathbb{G}_{\mathcal{S}_t} + \beta_t d^{t-1}, x^{t+1} = x^t - \eta_t \left[ (1 - \nu_t) \mathbb{G}_{\mathcal{S}_t} + \nu_t d^t \right] \)
where the parameter $\nu_t \in [0, 1]$ interpolates between SGD ~\citep{robbins1951stochastic}($\nu_t = 0$) and (normalized) SHB\citep{polyak1964some} ($\nu_t = 1$). When the parameters $\eta_t$, $\beta_t$ and $\nu_t$ are held constant (thus the subscript $t$ can be omitted) and $\nu = \beta$, it recovers a normalized variant of NAG~\citep{nesterov2018} with an additional coefficient $1 - \beta_t$ on the stochastic gradient term. For adaptive learning rate methods (e.g., Adam\(\left(\phi(\mathbb{G}_{\mathcal{S}_t}) = \sqrt{\beta_2 \mathbb{G}_{\mathcal{S}_t}^2 + (1 - \beta_2) v_{t-1}} \right)\)~\citep{kingma2014adam}, RMSProp~\citep{hinton2012neural}, Adagrad~\citep{duchi2011adaptive}, etc.), we simply abstract them as \(\eta_t^\phi = \frac{\eta_t}{\phi(\mathbb{G}_{\mathcal{S}_t})}\).

\begin{theorem} \label{theorem:SA} (SA) Assuming the clients satisfy Assumptions \ref{assumption:local_optimal_heter} and \ref{assumption:eff_dec} ,and the single-round update distance is: 
\begin{equation}
\left\| x_{t+1} - x^*_{\mathcal{S}_t} \right\|^2  = \left(1 - \wp_{\nu,\beta}^{\phi,t}  \right) \left\| \delta_{\mathcal{S}_t}^t \right\|^2  - \eta_t^\phi \left( \eta_t^\phi \left( 1 - \nu_t \beta_t \right) \Delta_{t+1} + \nu_t \beta_t  \eth_\phi \left\|  d^{t-1}  \right\| \right),
\end{equation}  where \( \wp_{\nu,\beta}^{\phi,t}  = 2 \hat{\eta}_{\nu,\beta}^{\phi,t} (1 - \hat{\eta}_{\nu,\beta}^{\phi,t}) ( 1 - \cos \langle \delta_{\mathcal{S}_t}^K , \delta_{\mathcal{S}_t} \rangle \sigma_\Delta  ) , \eth_\phi = [ \eta_{\nu,\beta}^{\phi,t} \left\|  d^{t-1} \right\| - 2 \hbar_{\phi} \left\| \delta_{\mathcal{S}_t}  \right\| ] , \hbar_\phi = ( (\hat{\eta}_{\nu,\beta}^{\phi,t} - 1 ) \cos \langle \delta_{\mathcal{S}_t} ,  d^{t-1} \rangle - \hat{\eta}_{\nu,\beta}^{\phi,t} \sigma_\Delta \cos \langle \delta_{\mathcal{S}_t}^K ,  d^{t-1} \rangle ) \) and \( \hat{\eta}_{\nu,\beta}^{\phi,t} =\eta_t^\phi \left( 1 - \nu_t \beta_t \right) , \eta_{\nu,\beta}^{\phi,t} = - \eta_t^\phi \nu_t \beta_t\).

\textit{Role of \(\eta^\phi_t,\nu_t,\beta_t\)} These parameters cause an overall decay to the right-hand side of the equation. 
Focusing on \(\eta_t^\phi ( 1 - \nu_t \beta_t ) \Delta_{t+1} + \nu_t \beta_t  \eth_\phi \|  d^{t-1}  \|\), under appropriate coefficient conditions, the decrease in distance can be viewed as a weighted sum of \(\Delta_{t+1}\) and \(d^{t-1}\), where the weights sum approx to 1. When \(A\) is indefinite, \(\Delta_{t+1}\) becomes ineffective, but \(d^{t-1}\) maybe can still act as a positive term that brings \(x_{t+1}\) closer to \(x^*_{\mathcal{S}_t}\). Therefore, when \( d^{t-1} \) satisfies the correction term condition, its role is to increase the update distance per step (which manifests as a certain acceleration effect over the entire round) and to correct the trajectory oscillations caused by the action of \(A\). Therefore, when \( d^{t-1} \) satisfies the correction term condition \ref{corollary:correct-term}, its role is to increase the update distance per step (which manifests as a certain acceleration effect over the entire round) and to correct the trajectory oscillatorys caused by the indefinite of \( A \).
\end{theorem}


\newpage

\appendix

\section{Related Work: Theoretical Perspectives on Gradient Heterogeneity}
\label{section:related work}
In federated learning (FL), data heterogeneity is often cited as the reason why standard convergence analyses—based on the local Lipschitz-gradient assumption and a gradient-divergence bound—predict deteriorating rates as the number of local steps grows. Yet empirical evidence shows that more local updates can actually accelerate training even when the product of the Lipschitz constant and divergence is large. To reconcile theory and practice, Wang et al.~\citep{wang2024new} replace the usual Lipschitz condition with a weaker \emph{heterogeneity-driven pseudo-Lipschitz} assumption. Under this and the standard divergence bound, they derive a tighter upper bound for FedAvg (and its variants), where the large local Lipschitz constant is replaced by a much smaller pseudo-Lipschitz constant, without altering the asymptotic order. It is shown in Fig. \ref{fig:emnist_all},\ref{fig:gru_all} that the performance of more of the following algorithms deteriorates with increasing heterogeneity. The environment of each of these methods is uniform.

\subsection{Adaptive Optimization, Server Momentum, and Sharpness-Aware Methods}
\begin{table}[htbp]
\centering
\caption{The function properties only satisfy the convergence conclusion of the latest method analysis of L-smooth. In this table, \(\mathcal{F} = F(x_0) - F(x^*)\).} 
\adjustbox{width=\textwidth}{
\begin{tabular}{llll}
\hline
Research & Corollary& Local Learning Rate \\ \hline
\makecell[l]{Heterogeneity-driven \\ Pseudo-Lipschitz \citep{wang2024new}}  & \(\mathcal{O} \left( \sqrt{\frac{\mathcal{F}L\vartheta^2}{TK|\mathcal{S}|}} + \frac{\zeta^2 + \vartheta^2/K}{TKN} \right).\) & \( \mathcal{O} \left( \frac{1}{\sqrt{TKN}} \right) \). \\

FedGM \citep{sun2024role} & \(\mathcal{O} \left( \sqrt{\frac{K}{Tm}} \right)\)& \(\mathcal{O} \left( \frac{1}{\sqrt{TK}} \right) \).  \\

FedAVG-M \citep{cheng2024momentum}          & \( \mathcal{O} \left( \sqrt{\frac{L\Delta \vartheta^2}{NKT}} + \frac{L\Delta}{T} \right) \)& \( \mathcal{O} \left( \frac{1}{\beta K \eta_tL^2 T} \right) \).  \\

FEDAS\citep{wang2024fadas}& \(\mathcal{O}\left(\frac{\sqrt{\mathcal{F}} \vartheta}{\sqrt{T K M}} + \frac{\sqrt{\mathcal{F}} \zeta}{\sqrt{T M}} + \frac{\mathcal{F}}{T} + \frac{\mathcal{F} G}{T \sqrt{M}} + \frac{\mathcal{F} \tau_{\max} \tau_{\text{avg}}}{T}\right).\) & \( \mathcal{O}\left( \frac{\sqrt{\mathcal{F}}}{\sqrt{T K (\vartheta^2 + K \zeta^2)}} \right). \) \\
FedAMS\citep{wang2022communication}         & \(\mathcal{O}\left(\frac{\sqrt{K}}{\sqrt{T|\mathcal{S}_t|}}\right)\).        & \(\mathcal{O} \left(\frac{1}{\sqrt{T}K}\right)\) \\
FedAdam\citep{reddi2020adaptive}            & \(\mathcal{O} \left( \frac{\mathcal{F}}{\sqrt{|\mathcal{S}| KT}} + \frac{2 \vartheta_l^2 L}{G^2 \sqrt{|\mathcal{S}| KT}} + \frac{\vartheta^2}{G KT} + \frac{\vartheta^2 L \sqrt{|\mathcal{S}|}}{G^2 \sqrt{K} T^{3/2}} \right)\).& \(\mathcal{O} \left(\frac{1}{K L \sqrt{T}} \right) \). \\ \hline
\end{tabular}
}
\end{table}

Adaptive optimizers and momentum have long improved centralized training, and recently, several works have adapted them to FL. Reddi et al.~\citep{reddi2020adaptive} introduce federated versions of Adagrad, Adam, and Yogi, proving convergence under nonconvex heterogeneity and showing significant empirical gains. Sun et al.~\citep{sun2024role} propose a unified server-momentum framework that supports stage-wise scheduling and asynchronous clients, with rigorous convergence guarantees in heterogeneous settings. Kim et al.~\citep{kim2023adaptive} develop Delta-SGD, which auto-tunes each client’s step size via its local smoothness, matching or exceeding centralized baselines without extra tuning. Fan et al. (FedLESAM)~\citep{fan2024locally} argue that local SAM perturbations misalign with global sharpness; instead, they estimate the global perturbation by differencing successive global models, achieving tighter bounds and faster convergence in practice.

\subsection{Regularization and Drift-Correction Methods}

In convex FL, regularization and control variates correct client drift and improve communication–computation trade-offs. SCAFFOLD~\citep{karimireddy2020scaffold} uses client-side control variates to eliminate drift, reducing communication rounds and leveraging data similarity for quadratic objectives. DANE, DANE+ \& FedRed~\citep{jiang2024federated} revisit the proximal-point method: DANE enjoys communication reduction under Hessian similarity; DANE+ and FedRed introduce a doubly-regularized drift-correction scheme that relaxes local solver accuracy while preserving communication complexity. ProxSkip~\citep{mishchenko2022proxskip} skips expensive proximal updates in most iterations, cutting prox calls from \(O(\kappa\log\frac1\varepsilon)\) to \(O(\sqrt\kappa\log\frac1\varepsilon)\) while keeping overall iteration complexity, outperforming FedAvg and SCAFFOLD under heterogeneity. S-DANE~\citep{jiang2024stabilized} augments DANE with auxiliary prox-centers to further relax local accuracy requirements, supporting stochastic solvers and partial participation, and admits adaptive line-search variants. FedAVGM~\citep{cheng2024momentum} integrates server momentum into FedAvg and SCAFFOLD, proving FedAvg converges with a constant learning rate and no bounded-heterogeneity assumption, and accelerating SCAFFOLD under partial participation, with momentum-based variance-reduction extensions.


\begin{figure}[!ht]
    \centering
    \includegraphics[width=1\linewidth]{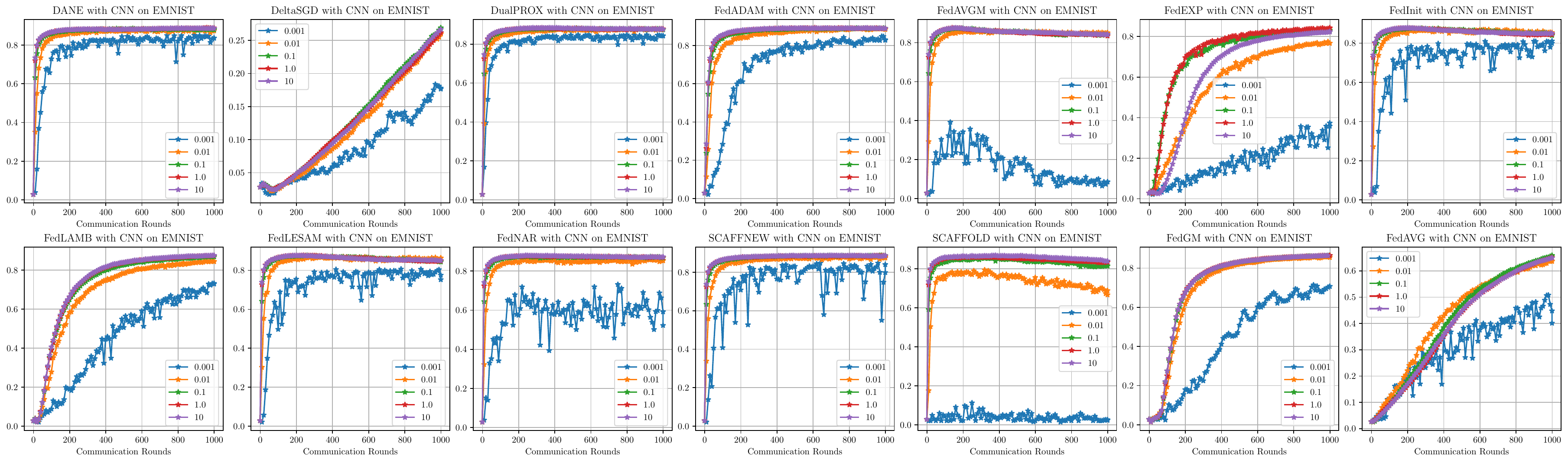}
    \caption{The performance of FedAVG, FedAVGM, DeltaSGD, DualPROX, FedEXP, FedGM, FedInit, FedLESAM, FedNAR, FedPROX, SCAFFOLD, SCAFFNEW, and FedADAM on the EMNIST classification task.} 
    \label{fig:emnist_all}
\end{figure}

\begin{figure}[!ht]
    \centering
    \includegraphics[width=1\linewidth]{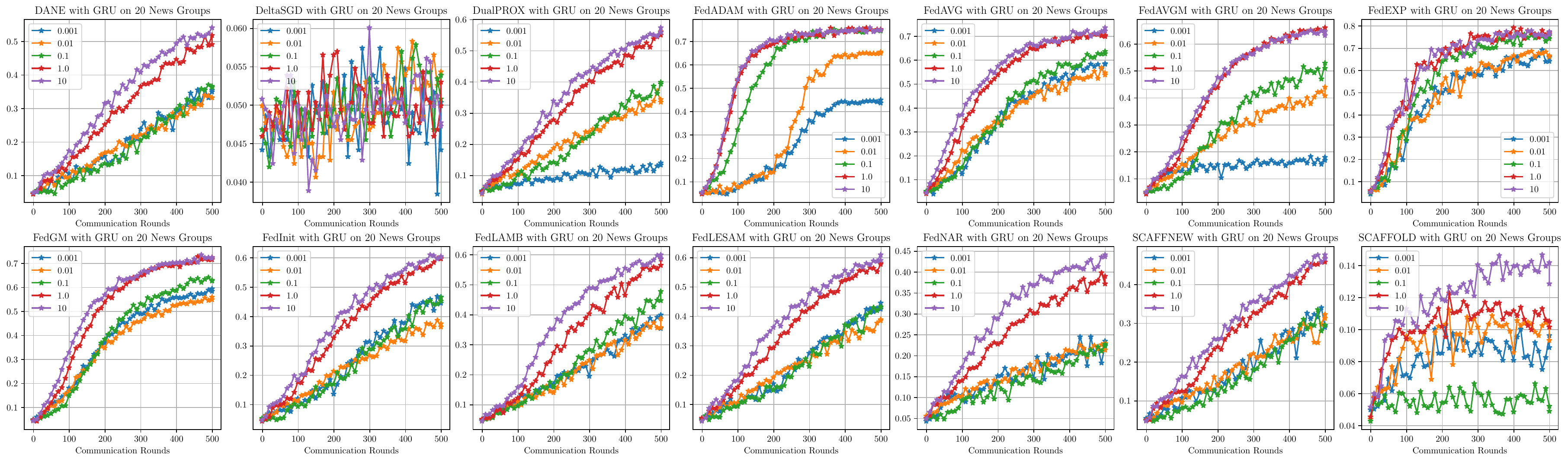}
    \caption{The performance of FedAVG, FedAVGM, DeltaSGD, DualPROX, FedEXP, FedGM, FedInit, FedLESAM, FedNAR, FedPROX, SCAFFOLD, SCAFFNEW, and FedADAM on the 20 News Groups classification task.}
    \label{fig:gru_all}
\end{figure}

\section{Visualization tools}
\label{section:Visualization Tools}

\begin{figure}
    \centering
    \includegraphics[width=1\linewidth]{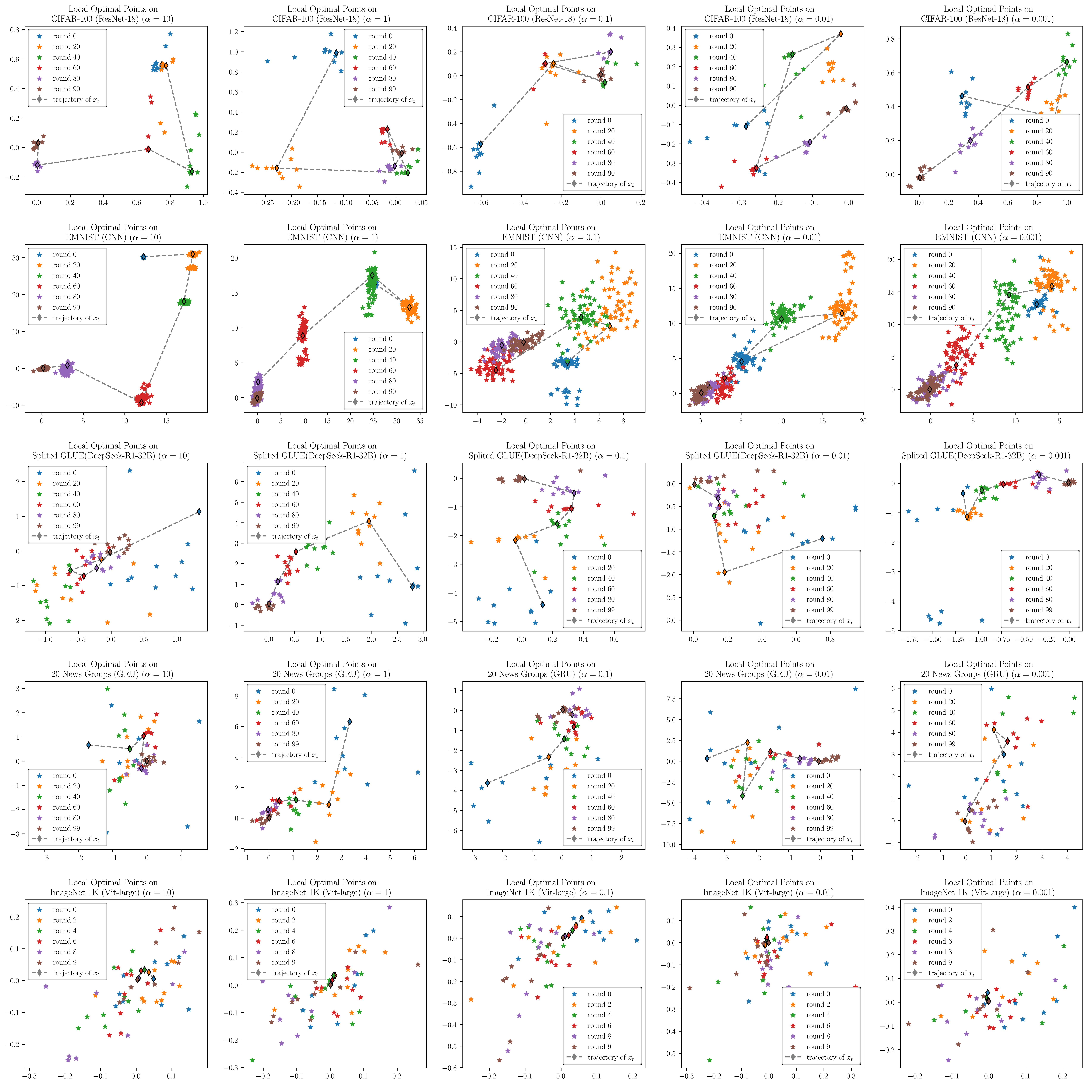}
    \caption{Visualization of the positional relationships between the optimization trajectory of \(x_t\) under multiple tasks and the local optimal points of different clients at time \(t\).} 
    \label{fig:traj_all}
\end{figure}

We refer to the method \citep{li2018visualizing} of visualizing loss landscapes, which has the advantage of being able to simultaneously visualize the relative regions of multiple neural networks. In this paper, our drawing includes the approximate positions of local optima in client-side neural networks and the trajectory of the client-side loss function. The method first initializes two "base models" \( W_x \), \( W_y \) using a Gaussian distribution as the basis for projection. This paper's visualization is divided into two cases:

\textbf{Relative Position Visualization Method (Fig. \ref{fig:assumption1_re} right) }: For locating model \( W_{i,t} \) in 2D coordinates \( x_i, y_i \). Extract its classification layer weights \( W_{i,t}^c \), and extract the classification layer weights of the base models \( W_x^c \) and \( W_y^c \). The coordinates are determined by projecting \( W_{i,t}^c \) onto \( W_x^c \) and \( W_y^c \), divided by the norm of the basis vectors:
\begin{equation}
x = \frac{ \| W_{i,t}^c \cos(\langle W_{i,t}^c, W_x^c \rangle) \| }{ \| W_x^c \| }, \quad y = \frac{ \| W_{i,t}^c \cos(\langle W_{i,t}^c, W_y^c \rangle) \| }{ \| W_y^c \| }
\label{eq:relateive_pos}
\end{equation} In Fig.~\ref{fig:traj_all}, we plotted the relative positions of optimal points and \( x_t \) across different tasks under varying communication rounds. Since the entire process shares a pair of common base models, the trajectory of \( x_t \) can be traced. The results in Fig.~\ref{fig:traj_all} reveal that in more non-i.i.d. scenarios, the positions of the optimal points become increasingly dispersed.


\textbf{Loss Landscape Visualization Method (Fig. \ref{fig:assumption1_re} left)}: All client-side models share the same set of base models \( W_x, W_y \). For a model \( W_{i,t} \) to be visualized on dataset \( D_i \), we plot the function \[ f(\sigma,\beta;x_i,y_i,\mathcal{D}_i) = f(W_{i,t} + (\sigma   - x_i)W_x + (\beta - y_i)W_y; D_i) \] where \(x_i,y_i\) are the relative positions of \(W_{i,t}\) calculated using Eq. \ref{eq:relateive_pos} and \( \sigma  , \beta \in \mathbb{R} \). Here, the addition and multiplication operations are applied to all parameters of the neural network. 

\textbf{Gather Local Optima Visualization} 
To gather all the functions near the client's optimal point on a single graph, we use the following form:
\(g(\sigma,\beta) =\\ \min (f(\sigma,\beta;\mathcal{D}_1), f(\sigma,\beta;\mathcal{D}_2), f(\sigma,\beta;\mathcal{D}_{\dots}))\) In Fig \ref{fig:assumption1_re},\ref{fig:gather_paraboloid} we use this expression. 


\section{Experiment Detail}
\label{section:Exp}
\subsection{Simulating the Experimental Setup for Heterogeneous} 

\subsubsection{Fixed Optimal Points Simulation: Paraboloid}

\begin{figure}[htbp]
  \centering
  \begin{minipage}[b]{0.5\textwidth}
    \centering
    \includegraphics[width=\linewidth]{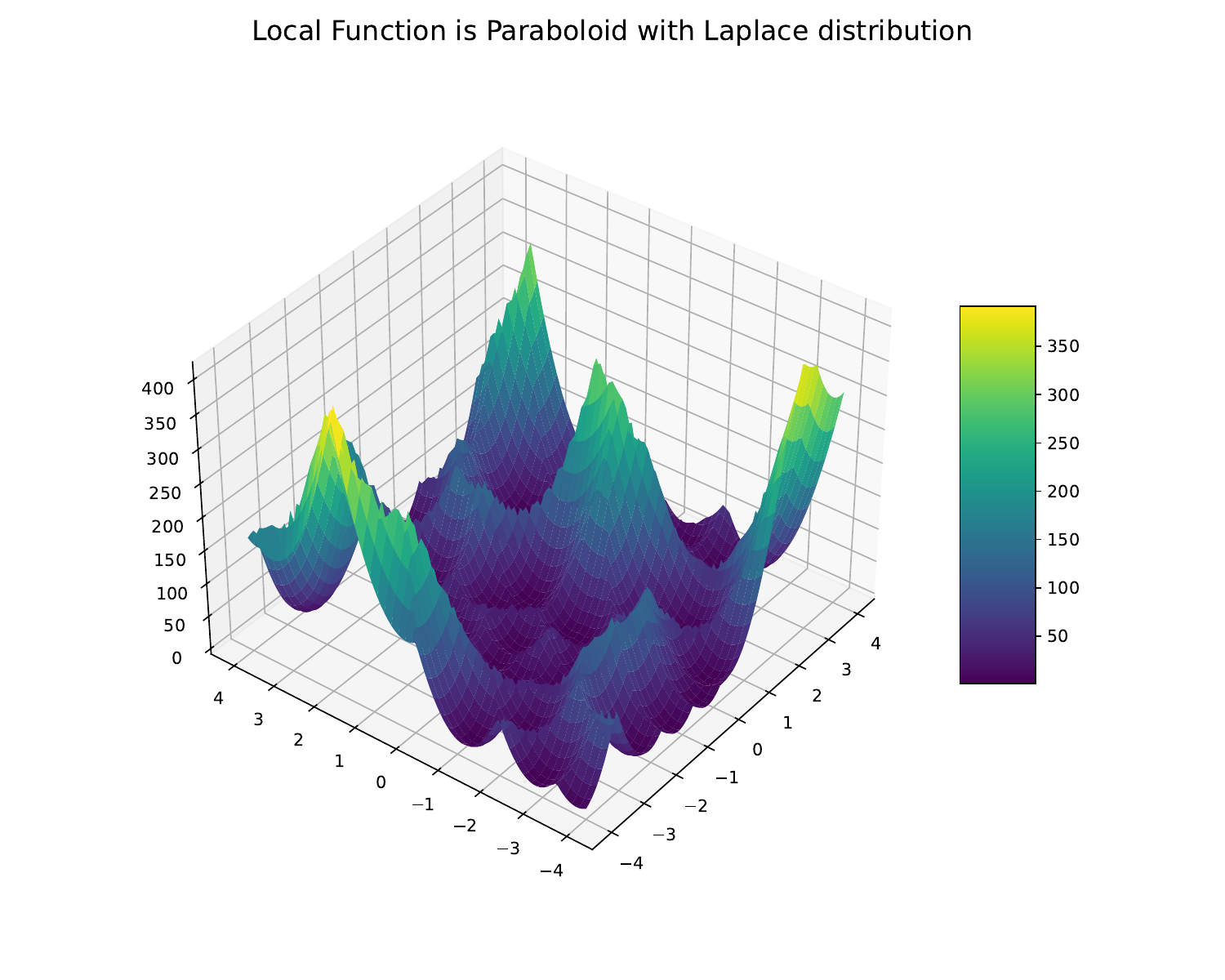}
  \end{minipage}
  \hfill
  \begin{minipage}[b]{0.49\textwidth}
    \centering
    \includegraphics[width=\linewidth]{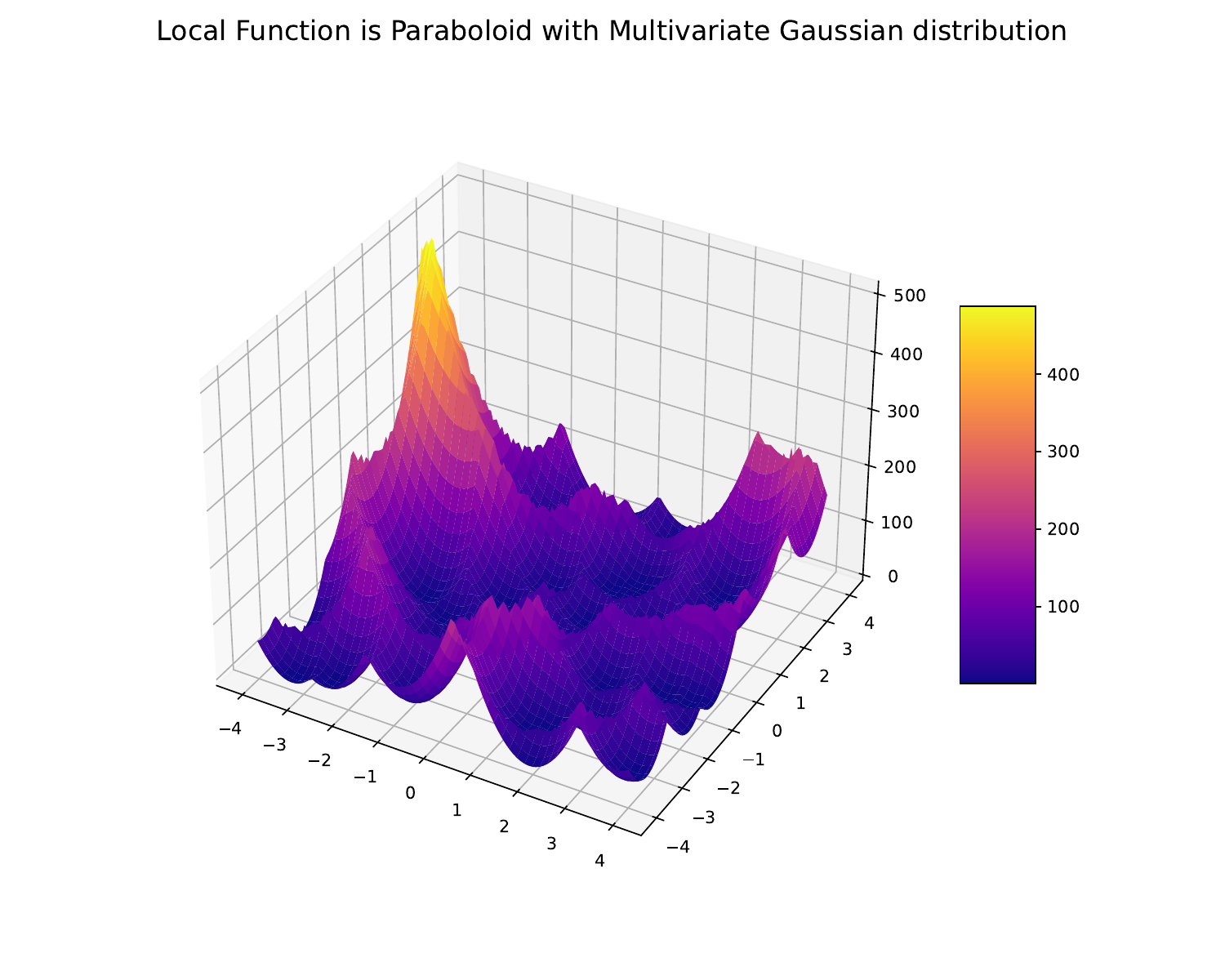}
  \end{minipage}
  \caption{When \(f_i\) is a parabolic surface, the merging situation near the optimal point on the client side.We use \(g(x) = \min (f_1(x), f_2(x), f_{\dots}(x))\) to display the optimal points for each client.} 
  \label{fig:gather_paraboloid}
\end{figure}

Since the optimal point of the neural network tends to change with the change of \(x_t\), in the simple verification of the existence of the oscillating region (Fig.~\ref{fig:main_theorem}), we use the fixed optimal point of the parabola to verify the oscillating region:
\[
 f(x) = \frac{1}{M} \sum_{i=0}^M (x - \delta_i)^2
\] where \(\delta_i\) samples from the multivariate Gaussian and Laplace distributions. In Fig. \ref{fig:gather_paraboloid} we plot the case of a localized image of \((x - \delta_i)^2\).

\subsection{Various neural network tasks}

\subsubsection{Methods for Data Splitting and Allocation}

We utilize the Dirichlet distribution to randomly allocate data across various classes detailed in Alg. \ref{alg:data_split}, thereby achieving a non-i.i.d. partitioning of data among multiple clients. First, for each class, the algorithm gathers the indices of all samples belonging to that class. Then, for each class, a probability vector is drawn from the Dirichlet distribution, reflecting the distribution proportion of data across the clients for that class. Next, by multiplying the sampled probability vector with the total number of remaining unassigned samples in the current class and taking the ceiling of the result, the number of samples to be assigned to each client for that class is obtained. To ensure randomness and fairness in the assignment process, the order in which the clients are processed is randomly permuted in each iteration.

After the data for all classes has been allocated, the algorithm performs a post-processing step to ensure that each client has a sufficient number of samples: if a client's sample count falls below a preset minimum threshold, some samples are transferred from the client with the most samples to meet the requirement. Finally, the algorithm outputs the set of data indices corresponding to each client, achieving a Dirichlet distribution-based non-i.i.d. data partitioning that effectively simulates the uneven distribution of data among clients in real-world scenarios.

\begin{algorithm}[ht]
\caption{Non-i.i.d Data Splitting via Dirichlet Allocation}
\textbf{Input}:Source data with labels $y$, number of clients $C$, Dirichlet parameter $\alpha  $. Data index allocation $\{D_1, D_2, \dots, D_C\}$ for each client.\\
\textbf{Partition:} For each class $c$, let $I_c \gets \{ i \mid y_i = c \}$.
\begin{algorithmic}[1] 
\FOR{classes $c$}
    \STATE Sample a probability vector $p^c \sim \text{Dirichlet}(\sigma  , \dots, \sigma  )$ over $C$ clients.
\ENDFOR
\WHILE{there exists a class $c$ with $|I_c| > 0$}
    \STATE Shuffle the client order: $\pi \gets$ random permutation of $\{1,\dots,C\}$.
    \FOR{classes $c$}
        \STATE Let $n_c \gets |I_c|$.
        \STATE Compute allocation counts: $d^c \gets \lceil p^c \cdot n_c \rceil$.
        \FOR{clients $i \in \pi$}
            \STATE Assign the first $d^c_i$ indices from $I_c$ to client $i$,
            \STATE i.e., update \( D_i \gets D_i \cup \text{first } d^c_i \text{ elements of } I_c.\)
            \STATE Remove these indices from $I_c$.
        \ENDFOR
    \ENDFOR
\ENDWHILE
\FOR{clients $i$}
    \IF{$|D_i|$ is below a minimal threshold}
        \STATE Transfer a few samples from the client with the most data.
    \ENDIF
\ENDFOR
\RETURN $\{D_1, D_2, \dots, D_C\}$.
\end{algorithmic}
\label{alg:data_split}
\end{algorithm}

\subsubsection{GRU on 20 News Groups}


The 20 Newsgroups\citep{Lang95} dataset is a widely used and well-known text classification dataset that contains articles from 20 different newsgroups, covering a wide variety of topics and categories. Our goal is to analyze the content features of each article and accurately predict its specific newsgroup category. 

In the data processing phase for the 20 Newsgroups dataset, text sequences are standardized to a fixed length of 128 tokens. For samples exceeding this length, a random truncation strategy is implemented: when the remaining sequence length after truncation exceeds 128 tokens, a random starting index between 0 and the remaining length is selected to preserve semantically critical segments. Shorter sequences are padded with zeros at the beginning to align all inputs to a fixed 128×1 tensor dimension. This approach enhances data randomness while maintaining batch processing efficiency, with zero-padded positions masked via the embedding layer's padding identifier (padding\_idx=0).

The network architecture employs a two-layer unidirectional GRU structure. The embedding layer maps a 128,000-dimensional discrete vocabulary into a 128-dimensional continuous vector space, followed by layer normalization to standardize the 128×128 sequence features. Each recurrent layer contains 128 hidden units, capturing temporal dependencies through time-step unfolding. The final hidden state at the 128th time step serves as the global feature representation. A fully connected layer projects the 128-dimensional hidden state into a 20-dimensional classification space, corresponding to the 20 newsgroup categories. The design utilizes parameter-sharing mechanisms to control complexity while maintaining temporal modeling capabilities. Layer normalization mitigates gradient vanishing issues, enhancing robustness for long-text processing.

\subsubsection{Resnet-18 on CIFAR 100}

The CIFAR-100\citep{krizhevsky2009learning} is a dataset for image classification tasks and is part of the CIFAR series of datasets.  FedOPT employs a two-step LDA process for both coarse and fine labels. We conduct training on CIFAR-100 using a modified ResNet-18 like FedOPT \citep{reddi2020adaptive}, wherein the batch normalization layers have been substituted with group normalization layers. Specifically, each group normalization layer is configured with two groups. The preprocessing for CIFAR-100 involves images with 3 channels of 32 $\times$ 32 pixels each, where each pixel is represented by an unsigned int8. We conduct preprocessing on both training and test images. For training images, random cropping is applied to reshape them to (24, 24, 3), followed by a random horizontal flip. For testing images, a central crop is performed to adjust the shape to (24, 24, 3).

\subsubsection{Vit-base for ImageNet 1k}

This experiment leverages a pre-trained Vision Transformer (ViT)~ \citep{radford2021learning} model to validate the effectiveness and linear separability of ViT representations in downstream classification tasks on the ImageNet-1K~\citep{imagenet15russakovsky} dataset. Key components include:

This experimental workflow consists of three core components. The dataset preparation phase utilizes the ImageNet-1K benchmark, with 1,281,167 training images and 50,000 validation images across 1,000 classes. Images undergo preprocessing through the ViTImageProcessor to meet the pretrained model specifications, including resizing to the required resolution, pixel value normalization to the [-1,1] range, and conversion into PyTorch tensors. 

For feature extraction, the pretrained CLIP-ViT-Large-Patch14 model – a Vision Transformer with a 14x14 patch division strategy – processes images in batches of 256 without gradient computation. The 768-dimensional [CLS] token embeddings from the final transformer layer (captured via last\_hidden\_state[:,0,:]) are aggregated into (N,768) feature matrices for both training and validation sets, accompanied by corresponding label vectors. The classification architecture employs a two-layer MLP implemented as torch.nn.Sequential: a 768→2048 linear projection followed by Tanh activation, then a 2048→1000 linear layer mapping to class logits. This classifier operates with frozen visual backbone parameters, focusing optimization exclusively on the final classification layer during fine-tuning.

\subsubsection{Deepseek-32b on Split GLUE}

In this experiment, we adopted multiple tasks from the GLUE\citep{wang2019glue} dataset and unified them into a 23-class multi-class classification task. Specifically, for each task (including CoLA, SST-2, MRPC, STS-B, QQP, MNLI, QNLI, RTE, and WNLI), we first defined the corresponding text fields and label fields based on the task configuration. For tasks with two text fields (e.g., MRPC, STS-B, MNLI, etc.), we concatenated the texts in the fields into a single input using predefined separators. Notably, for the STS-B task, since the original labels were continuous values, we converted them to integers via rounding to facilitate subsequent classification processing.

To ensure that labels from different tasks do not overlap with each other, we implemented a label offset (offset) strategy. Specifically, when processing each task, we added an accumulated offset to its label values, ensuring that the label distributions across tasks were non-overlapping and enabling the division of 23 distinct categories. Finally, we stored the processed training and validation sets from all tasks in a single file for subsequent model training and evaluation.

DeepSeek-R1~\citep{guo2025deepseek} is an advanced language model combining reinforcement learning (RL) and distillation to improve reasoning quality over predecessors. It uses two-stage RL (pattern exploration + human preference alignment) with supervised fine-tuning (SFT) and distills complex reasoning into a 32B-parameter Qwen-based architecture. The model excels in math, coding, and logic tasks, outperforms peers like GPT-4-mini in key areas, and supports advanced reasoning features (chain-of-thought, self-verification). Open-sourced for community-driven optimization and lightweight model distillation.

Using DeepSeek-R1-Distill-Qwen-32B as a feature extractor, we encode text through its pre-trained model to generate 5120-dimensional text representation vectors as input. The classification model employs a simplified multilayer perceptron (MLP) structure: the input layer projects 5120-dimensional features to a 1024-dimensional hidden layer via linear transformation, enhances non-linear expressive capacity through ReLU activation, and then directly outputs 23-dimensional logits (corresponding to 23 classes) via a second linear transformation.

\subsection{Total experimental environment setup}

\begin{table}[!ht]
  \centering
  \caption{Cross-device Configurations in Federated Learning Experimental Setups}
  \resizebox{\textwidth}{!}{%
    \begin{tabular}{@{}lccccccc@{}}
      \toprule
      \textbf{Model} & \textbf{Clients} & \textbf{Epochs} & \(\eta_l\) & \textbf{Batch Size}
        & \textbf{Comm.\ Rounds} & \textbf{Participants} & \textbf{Server LR} \\
      \midrule
      CNN on EMNIST  & 620   & 1 & 0.10  & 2048 & 1000 & 62 & 0.001 \\
      GRU on 20 News Groups & 100   & 1 & 0.001 & 512  & 500  & 10 & 0.010 \\
      ResNet-18 on CIFAR100 & 1000  & 5 & 0.001 & 128  & 4000 & 10 & 0.001 \\
      ViT-Large on ImageNet-1K            & 10000 & 5 & 0.001 & 768  & 1000 & 10 & 0.001 \\
      DeepSeek-R1-Distill-Qwen-32B on GLUE & 100   & 1 & 0.001 & 256  & 500  & 10 & 0.001 \\
      \bottomrule
    \end{tabular}%
  }
  \label{tab:fl_setups}
\end{table}

This paper focuses on the application of Federated Learning in a cross-device scenario \citep{liu2024vertical}, where clients such as smartphones \citep{yang2019federated} are reluctant to upload local private data, while FL enables clients to participate in the training process by transmitting locally trained models. The experiments specify the experimental conditions for Fig.~\ref{fig:acc_heter}, \ref{fig:emnist_all}, and \ref{fig:gru_all} in Tab.~\ref{tab:fl_setups}, which align with the cross-device scenario. Here, the Server Learning Rate (Server LR \(\eta\)) corresponds to the \(\eta\) parameter in the FedAdam algorithm. In the last few dozen rounds of the communication in Fig.~\ref{fig:acc_heter}, we will gradually reduce \(\eta_l\) to converge to a local optimum. Our experiments were conducted on a system running Ubuntu 24.04.2 LTS, equipped with 8 NVIDIA RTX 3090 GPUs and 512GB of memory.


\section{Proofs}
\label{section:proofs}

\subsection{Theorem and Corollary of LA-FedAVG}

\begin{theorem} 
(LA-FedAVG) Assuming that local optimal points satisfy Assumption \ref{assumption:local_optimal_heter} and all clients \( i \in \mathcal{S}_t \) satisfy the effective descent condition~\ref{assumption:eff_dec}.  The single update distance from \( x_{t+1} \) of LA-FedAVG  to the sampled local optimum \( x^*_{\mathcal{S}_t} = \sum_{i \in \mathcal{S}_t} \rho_i x_i^* \) is:
\begin{equation}
\label{eq:main_prf}
\left\| x_{t+1} - x^*_{\mathcal{S}_t} \right\|^2  = \left\| x_{t} - x^*_{\mathcal{S}_t} \right\|^2  - \frac{1}{|\mathcal{S}_t|^2} \left(\mathcal{P}_{\mathcal{S}_t} \cdot X_{\mathcal{S}_t} \right)^\top A  \left(\mathcal{P}_{\mathcal{S}_t} \cdot X_{\mathcal{S}_t} \right),
\end{equation}
where \(\mathcal{P}_{\mathcal{S}_t} = \left[\rho_i^t\right]_{i \in \mathcal{S}_t} \) is weight vector, \(X_{\mathcal{S}_t} = \left[ \delta_{t,i} \right]_{i \in \mathcal{S}_t} \), the element of matrix A  is \( A_{i,j} = \cos \left\langle  \delta_{t,i},  \delta_{t,j} \right\rangle - \sigma_i^t \sigma_j^t  \cos \left\langle  \delta_{t,i}^K, \delta_{t,j}^K \right\rangle \). 

\begin{proof}
    With any \(\sum_{i \in \mathcal{S}_t} \rho_i^t = 1 , \forall \rho_i^t > 0 , x^*_{t,F} = \sum_{i \in \mathcal{S}_t} \rho_i^t x_i^*  \) 

\begin{align*}
\left\| x_{t+1} - x^*_{t,F} \right\|^2 &= \left\| \sum_{i \in \mathcal{S}_t} \rho_i^t \left( x_{K,i}^t  - x^*_i \right) \right\| = \left\| \sum_{i \in \mathcal{S}_t} \rho_i^t \delta_{t,i}^K \right\|^2 = \frac{1}{|\mathcal{S}_t|^2} \left[ \sum_{i\in \mathcal{S}_t}\sum_{j\in \mathcal{S}_t} \left\langle \rho_i^t \delta_{t,i}^K,\rho_j^t \delta_{t,j}^K \right\rangle   \right]  \\
&= \frac{1}{|\mathcal{S}_t|^2} \left[ \sum_{i\in \mathcal{S}_t}\sum_{j\in \mathcal{S}_t} \left\| \rho_i^t \delta_{t,i}^K  \right\| \left\| \rho_j^t \delta_{t,j}^K \right\| \cos \left\langle \rho_i^t \delta_{t,i}^K,\rho_j^t \delta_{t,j}^K \right\rangle   \right] 
\end{align*}

For the second inequality, we let \( \delta_{t,i}^K = x_{K,i}^t - x_i^* \). The second last inequality uses the vector cosine similarity formula: the cosine of the angle 
between vectors a and b is given by their dot product divided by the product of their magnitudes.

\[
\left \langle \mathbf{a} \cdot\mathbf{b}  \right\rangle  =   \|\mathbf{a}\| \|\mathbf{b}\| \cos \left \langle \mathbf{a} \cdot\mathbf{b} \right\rangle 
\]

Similarly, we can express the distance between \( x_t \) and \( x_{t,F}^* \) in the following form:

\begin{align*}
\left \| x_t - x^*_{t,F} \right\|^2 &= \left\| \sum_{i \in \mathcal{S}_t} \rho_i^t \left( x^t  - x^*_i \right) \right\|^2 = \frac{1}{|\mathcal{S}_t|^2} \left[ \sum_{i\in \mathcal{S}_t}\sum_{j\in \mathcal{S}_t} \left\| \rho_i^t \delta_{t,i}  \right\| \left\| \rho_j^t \delta_{t,j} \right\| \cos \left\langle \rho_i^t \delta_{t,i},\rho_j^t \delta_{t,j} \right\rangle   \right] 
\end{align*}

Let \( \delta_{t,i} = x^t - x_i^* \).We need to determine under what conditions \( x_{t+1} \) will converge toward \( x_{t,F}^* \), so
by subtracting the above two equations, we obtain:

\begin{align*}
&\left\| x_{t+1} - x^*_{t,F} \right\|^2 - \left\| x_{t} - x^*_{t,F} \right\|^2 \\
=& \frac{1}{|\mathcal{S}_t|^2} \left[ \sum_{i\in \mathcal{S}_t}\sum_{j\in \mathcal{S}_t} \left( \left\| \rho_i^t \delta_{t,i}^K  \right\| \left\| \rho_j^t \delta_{t,j}^K \right\| \cos \left\langle \rho_i^t \delta_{t,i}^K,\rho_j^t \delta_{t,j}^K \right\rangle  - \left\| \rho_i^t \delta_{t,i}  \right\| \left\| \rho_j^t \delta_{t,j} \right\| \cos \left\langle \rho_i^t \delta_{t,i},\rho_j^t \delta_{t,j} \right\rangle \right) \right] \\
=& \frac{1}{|\mathcal{S}_t|^2} \left[ \sum_{i\in \mathcal{S}_t}\sum_{ j\in \mathcal{S}_t }  \rho_i^t \rho_j^t \left\| \delta_{t,i}  \right\| \left\| \delta_{t,j} \right\| \left( \sigma_i^t \sigma_j^t  \cos \left\langle  \delta_{t,i}^K, \delta_{t,j}^K \right\rangle - \cos \left\langle  \delta_{t,i},  \delta_{t,j} \right\rangle \right) \right]
\end{align*}
\end{proof}

\end{theorem}

\begin{lemma} Decoupling \citep{vershynin2011simple,vershynin2009high,FR13} is a technique of replacing quadratic forms of random variables by bilinear forms. Let \(A\) be an \(n \times n\) matrix with zero diagonal. Let \(X = (X_1,...,X_n)\) be  a random matrix.
\begin{equation}
\label{lemma:decoupling}
\mathbb{E} \sum_{i \in [n]} \sum_{j \in [n]} a_{ij} \left\langle X_i , X_j \right\rangle = 4 \mathbb{E} \sum_{i \in I} \sum_{\substack{j \in I^c \\ j \ne i}} a_{ij} \left\langle X_i , X_j^{'} \right\rangle
\end{equation}

where \(X'\) is an independent copy of \(X\) , \(  I := \{i : \delta_i = 1\} \text{~and~} I^c := [n] / I \).

\begin{proof}
We adopted some proof ideas from the research to address the chaos in quadratic forms. Study point \citep{vershynin2011simple,vershynin2009high,FR13} that they replace the chaos \( \sum_{i,j} a_{ij} \left\langle X_i , X_j \right\rangle\) by the ``partial chaos''
\[
\sum_{(i,j) \in I \times I^c} a_{ij} \left\langle X_i , X_j \right\rangle
\]
where the subset of indices \(I \subset \{1, \ldots, n\}\) will be chosen by random sampling. The advantage of partial chaos is that the summation is done over disjoint sets for \(i\) and \(j\). Thus one can automatically replace \(X_j\) by \(X_j'\) without changing the distribution. Finally, the study completes the partial chaos to the full sum \( \sum_{i,j} a_{ij} 
 \left\langle X_i , X_j' \right\rangle \). 

To randomly select a subset of indices \( I \), let us consider selectors of column \( \delta_1, \ldots, \delta_n \in \{0, 1\} \), which are independent Bernoulli random variables with \( \mathbb{P}\{\delta_i = 0\} = \mathbb{P}\{\delta_i = 1\} = 1/2 \). Define

\[ I := \{i : \delta_i = 1\}. \]

Condition on \( X \). Since by assumption \( a_{ii} = 0 \) and

\[ \mathbb{E} \delta_i (1 - \delta_j) = \frac{1}{2} \cdot \frac{1}{2} = \frac{1}{4} \quad \text{for all } i \neq j, \]

we may express the chaos as

\[ \sum_{i \in [n]} \sum_{\substack{j \in [n] \\ i \neq j}} a_{ij}\left\langle X_i , X_j \right\rangle = 4 \mathbb{E}_\delta  \sum_{i \ne j} \delta_i (1 - \delta_j) a_{ij} \left\langle X_i , X_j \right\rangle = 4 \mathbb{E}_I \sum_{i \in I} \sum_{i \in I^c} a_{ij} \left\langle X_i , X_j \right\rangle .\]

Since \(X_i, \, i \in I\) are independent from \(X_j, \, j \in I^c\), the distribution of this sum will not change if we replace \(X_j\) by \(X'_j\), the coordinates of \(X'\), and thus we can obtain Lemma \ref{lemma:decoupling}.

\end{proof}

\end{lemma}

\begin{corollary} Under the conditions outlined in Theorem \ref{theorem:main}, assuming that \( x_{i,t}^K \) is uniformly distributed within a norm ball centered at \( x_i^* \) with radius \( ||\delta_i^t|| \), and given that \( \min_{i,j \in \mathcal{S}_t} \cos\langle \delta_{t,i}, \delta_{t,j} \rangle > 0 \) (which implies that the matrix \( A \) is positive definite), through decoupling Lemma \ref{lemma:decoupling}, we can determine the range of the descent distance \(\Delta_{t+1} =\\ \frac{1}{|\mathcal{S}_t|^2} \left(\mathcal{P}_{\mathcal{S}_t} \cdot X_{\mathcal{S}_t} \right)^\top A  \left(\mathcal{P}_{\mathcal{S}_t} \cdot X_{\mathcal{S}_t} \right)\) in the LA-FedAVG algorithm as follows:
\begin{equation}
\rho_{min,t}^2 \mho_{min} \left\| \delta^{t,i}_{min}  \right\|^2  \le \mathbb{E}  \left[ \Delta_{t+1} \right] \le \rho^2_{max,t} \mho_{max} \left\| \delta^{t,i}_{max}  \right\|^2,
\end{equation}
where \( \mho_{max} = \frac{1}{|\mathcal{S}_t|} (1 - \sigma_{min,t}^2 ) +  \max_{i,j \in \mathcal{S}_t} \cos \left\langle \delta_{t,i}, \delta_{t,j} \right\rangle \) and \( \mho_{min} =  \frac{1}{|\mathcal{S}_t|} ( 1 - \sigma_{max,t}^2 ) + 4 \epsilon (1 - \epsilon) \min_{i,j \in \mathcal{S}_t } \cos \left\langle  \delta_{t,i}, \delta_{t,j} \right\rangle \) and \(\epsilon \in (0,1)\) is a constant due to the sampling bias.

\begin{proof}

\begin{equation}
\resizebox{\textwidth}{!}{$
\begin{aligned}
&\Delta_{t+1} = \frac{1}{|\mathcal{S}_t|^2} \left[ \sum_{i\in \mathcal{S}_t}\sum_{ j\in \mathcal{S}_t }  \rho_i^t \rho_j^t \left\| \delta_{t,i}  \right\| \left\| \delta_{t,j} \right\| \left( \sigma_i^t \sigma_j^t  \cos \left\langle  \delta_{t,i}^K, \delta_{t,j}^K \right\rangle - \cos \left\langle  \delta_{t,i},  \delta_{t,j} \right\rangle \right) \right] \\
&= \frac{1}{|\mathcal{S}_t|^2} \left[ \sum_{i\in \mathcal{S}_t}\sum_{ \substack{ j\in \mathcal{S}_t \\ i \ne j }}  \rho_i^t \rho_j^t \left\| \delta_{t,i}  \right\| \left\| \delta_{t,j} \right\| \left( \sigma_i^t \sigma_j^t  \cos \left\langle  \delta_{t,i}^K, \delta_{t,j}^K \right\rangle - \cos \left\langle  \delta_{t,i},  \delta_{t,j} \right\rangle \right) + \sum_{i \in \mathcal{S}_t } \rho_{i,t}^2 (\sigma_{i,t}^2 - 1) \left\| \delta_{t,i}  \right\|^2  \right] \\
&\le \frac{1}{|\mathcal{S}_t|^2} \left[ \sum_{i\in \mathcal{S}_t}\sum_{ \substack{ j\in \mathcal{S}_t \\ i \ne j }}  \rho_i^t \rho_j^t \left\| \delta_{t,i}  \right\| \left\| \delta_{t,j} \right\| \left( \sigma_i^t \sigma_j^t  \cos \left\langle  \delta_{t,i}^K, \delta_{t,j}^K \right\rangle - \min_{i,j \in \mathcal{S}_t } \cos \left\langle  \delta_{t,i},  \delta_{t,j} \right\rangle \right) + \sum_{i \in \mathcal{S}_t } \rho_{i,t}^2 (\sigma_{i,t}^2 - 1) \left\| \delta_{t,i}  \right\|^2  \right] \\
\end{aligned}
$} \nonumber
\end{equation}
To analyze the expectation of the first term within the square brackets, we need to decouple \( \sigma_i^t \sigma_j^t \cos \left\langle  \delta_{t,i}^K, \delta_{t,j}^K \right\rangle \). By decoupling~\ref{lemma:decoupling} the cosine values of pairs of vectors, they become independent, which facilitates the calculation of the expectation for each term in the summation.

\begin{equation}
\resizebox{\textwidth}{!}{$
\begin{aligned}
& \mathbb{E}  \left[ \Delta_{t+1} \right] \\
&\le \frac{1}{|\mathcal{S}_t|^2}  \left[ 4 \mathbb{E} \sum_{i \in I}\sum_{ \substack{ j\in I^c \\ i \ne j }}  \rho_i^t \rho_j^t \left\| \delta_{t,i}  \right\| \left\| \delta_{t,j}' \right\| \left( \sigma_i^t \sigma_j^t  \cos \left\langle  \delta_{t,i}^K, \delta_{t,j,K}' \right\rangle - \min_{i,j \in \mathcal{S}_t } \cos \left\langle  \delta_{t,i},  \delta_{t,j} \right\rangle \right) + \sum_{i \in \mathcal{S}_t } \rho_{i,t}^2 (\sigma_{i,t}^2 - 1) \left\| \delta_{t,i}  \right\|^2  \right] \\
&= \frac{1}{|\mathcal{S}_t|^2}  \left[ 4 \sum_{i \in I}\sum_{ \substack{ j\in I^c \\ i \ne j }}  \rho_i^t \rho_j^t \left\| \delta_{t,i}  \right\| \left\| \delta_{t,j}' \right\| \left( \sigma_i^t \sigma_j^t   \mathbb{E} \left[ \cos \left\langle  \delta_{t,i}^K, \delta_{t,j,K}' \right\rangle \right] - \min_{i,j \in \mathcal{S}_t } \cos \left\langle  \delta_{t,i},  \delta_{t,j} \right\rangle \right) + \sum_{i \in \mathcal{S}_t } \rho_{i,t}^2 (\sigma_{i,t}^2 - 1) \left\| \delta_{t,i}  \right\|^2  \right] \\
&= \frac{1}{|\mathcal{S}_t|^2}  \left[ 4 \sum_{i \in I}\sum_{ \substack{ j\in I^c \\ i \ne j }}  \rho_i^t \rho_j^t \left\| \delta_{t,i}  \right\| \left\| \delta_{t,j}' \right\| \left(- \min_{i,j \in \mathcal{S}_t } \cos \left\langle  \delta_{t,i},  \delta_{t,j} \right\rangle \right) + \sum_{i \in \mathcal{S}_t } \rho_{i,t}^2 (\sigma_{i,t}^2 - 1) \left\| \delta_{t,i}  \right\|^2  \right] \\
\end{aligned}
$} \nonumber
\end{equation}  where \( I := \{ i : \delta_i = 1 \} \), \( I^c := \mathcal{S}_t / I \), and \( \delta_{t,j}' \), \( \delta_{t,j,K}' \) denote independent copies of \( \delta_{t,j} \) and \( \delta_{t,j,K} \). Given that the inner product terms \( \delta_{t,i}^K, \delta_{t,j,K}' \) are independent and assuming that the angles between vectors are uniformly distributed, since the cosine of these angles is symmetric over the interval \([0, \pi]\), their integral—and thus their expectation—is zero, \(\mathbb{E} \left[ \cos \left\langle  \delta_{t,i}^K, \delta_{t,j,K}' \right\rangle \right] =  \int_{0}^{\pi}  \cos \theta \mathrm{d} \theta = 0 \).

Under the condition that client update directions are independent and their angles are uniformly distributed, the single-update absolute descent in federated learning \(\mathbb{E} [\Delta_{t+1}] < 0\) is guaranteed if and only if \( \min_{i,j \in \mathcal{S}_t} \cos\langle \delta_{t,i}, \delta_{t,j} \rangle > 0 \).

If \( \min_{i,j \in \mathcal{S}_t} \cos\langle \delta_{t,i}, \delta_{t,j} \rangle > 0 \) , we got :
\begin{align*}
\mathbb{E}  \left[ \Delta_{t+1} \right] &\le \frac{\min_{i \in \mathcal{S}_t}  \left\| \delta_{t,i}  \right\|^2   }{|\mathcal{S}_t|^2}  \left[ 4 \sum_{i \in I}\sum_{ \substack{ j\in I^c \\ i \ne j }}  \rho_i^t \rho_j^t  \left(- \min_{i,j \in \mathcal{S}_t } \cos \left\langle  \delta_{t,i},  \delta_{t,j} \right\rangle \right) + \sum_{i \in \mathcal{S}_t } \rho_{i,t}^2 (\sigma_{i,t}^2 - 1) \right] \\
&\le \frac{\min_{i \in \mathcal{S}_t}  \left\| \delta_{t,i}  \right\|^2   }{|\mathcal{S}_t|^2}  \left[ - 4 I I^c \rho_{t,min}^2 \min_{i,j \in \mathcal{S}_t } \cos \left\langle  \delta_{t,i},  \delta_{t,j} \right\rangle + \left| \mathcal{S}_t \right| \rho_{min,t}^2 (\sigma_{max,t}^2 - 1) \right] \\
&\le - \frac{  \rho_{min,t}^2   }{|\mathcal{S}_t|}  \left[ 1 + 4 \frac{I I^c}{|\mathcal{S}_t|}  \min_{i,j \in \mathcal{S}_t } \cos \left\langle  \delta_{t,i},  \delta_{t,j} \right\rangle - \sigma_{max,t}^2 \right] \times \left\| \delta^{t,i}_{min}  \right\|^2  \\
&\le - \rho_{min,t}^2  \left[ \frac{1}{|\mathcal{S}_t|} ( 1 - \sigma_{max,t}^2 ) + 4 \epsilon (1 - \epsilon )  \min_{i,j \in \mathcal{S}_t } \cos \left\langle  \delta_{t,i},  \delta_{t,j} \right\rangle  \right] \times \left\| \delta^{t,i}_{min}  \right\|^2  
\end{align*} where \(\epsilon = I / \left| 
\mathcal{S}_t \right|\) .

Next we proof inf of \(\Delta_{t+1}\)  
\begin{equation}
\resizebox{\textwidth}{!}{$
\begin{aligned}
&\Delta_{t+1} = \frac{1}{|\mathcal{S}_t|^2} \left[ \sum_{i\in \mathcal{S}_t}\sum_{ j\in \mathcal{S}_t }  \rho_i^t \rho_j^t \left\| \delta_{t,i}  \right\| \left\| \delta_{t,j} \right\| \left( \sigma_i^t \sigma_j^t  \cos \left\langle  \delta_{t,i}^K, \delta_{t,j}^K \right\rangle - \cos \left\langle  \delta_{t,i},  \delta_{t,j} \right\rangle \right) \right] \\
&= \frac{1}{|\mathcal{S}_t|^2} \left[ \sum_{i\in \mathcal{S}_t}\sum_{ \substack{ j\in \mathcal{S}_t \\ i \ne j }}  \rho_i^t \rho_j^t \left\| \delta_{t,i}  \right\| \left\| \delta_{t,j} \right\| \left( \sigma_i^t \sigma_j^t  \cos \left\langle  \delta_{t,i}^K, \delta_{t,j}^K \right\rangle - \cos \left\langle  \delta_{t,i},  \delta_{t,j} \right\rangle \right) + \sum_{i \in \mathcal{S}_t } \rho_{i,t}^2 (\sigma_{i,t}^2 - 1) \left\| \delta_{t,i}  \right\|^2  \right] \\
&\ge \frac{1}{|\mathcal{S}_t|^2} \left[ \sum_{i\in \mathcal{S}_t}\sum_{ \substack{ j\in \mathcal{S}_t \\ i \ne j }}  \rho_i^t \rho_j^t \left\| \delta_{t,i}  \right\| \left\| \delta_{t,j} \right\| \left( \sigma_i^t \sigma_j^t  \cos \left\langle  \delta_{t,i}^K, \delta_{t,j}^K \right\rangle - \max_{i,j \in \mathcal{S}_t} \cos \left\langle  \delta_{t,i},  \delta_{t,j} \right\rangle \right) + \sum_{i \in \mathcal{S}_t } \rho_{i,t}^2 (\sigma_{i,t}^2 - 1) \left\| \delta_{t,i}  \right\|^2  \right] \\
\end{aligned}
$} \nonumber
\end{equation} 

\begin{align*}
\mathbb{E}  \left[ \Delta_{t+1} \right] \ge \frac{1}{|\mathcal{S}_t|^2} \left[ 4 \sum_{i \in I}\sum_{ \substack{ j\in I^c \\ i \ne j }}  \rho_i^t \rho_j^t \left\| \delta_{t,i}  \right\| \left\| \delta_{t,j} \right\| \left(- \max_{i,j \in \mathcal{S}_t} \cos \left\langle  \delta_{t,i},  \delta_{t,j} \right\rangle \right) + \sum_{i \in \mathcal{S}_t } \rho_{i,t}^2 (\sigma_{i,t}^2 - 1) \left\| \delta_{t,i}  \right\|^2  \right]
\end{align*}
If \(\max_{i,j \in \mathcal{S}_t} \cos \left\langle  \delta_{t,i},  \delta_{t,j} \right\rangle > 0 \) , we get :
\begin{align*}
\mathbb{E}  \left[ \Delta_{t+1} \right] &\ge \frac{\max_{i \in \mathcal{S}_t} \left\| \delta_{t,i}  \right\| }{|\mathcal{S}_t|^2} \left[ 4 \sum_{i \in I}\sum_{ \substack{ j\in I^c \\ i \ne j }}  \rho_i^t \rho_j^t  \left(- \max_{i,j \in \mathcal{S}_t} \cos \left\langle  \delta_{t,i},  \delta_{t,j} \right\rangle \right) + \sum_{i \in \mathcal{S}_t } \rho_{i,t}^2 (\sigma_{i,t}^2 - 1) \right] \\
&\ge \frac{\rho^2_{max,t} } {|\mathcal{S}_t|^2} \left[  - 4  I I^c \max_{i,j \in \mathcal{S}_t} \cos \left\langle  \delta_{t,i}, \delta_{t,j} \right\rangle + \sum_{i \in \mathcal{S}_t } (\sigma_{min,t}^2 - 1) \right] \times \left\| \delta^{t,i}_{max}  \right\|^2 \\
&\ge \frac{ \rho^2_{max,t} } {|\mathcal{S}_t|^2} \left[ - 4  I I^c \max_{i,j \in \mathcal{S}_t} \cos \left\langle  \delta_{t,i}, \delta_{t,j} \right\rangle + \sum_{i \in \mathcal{S}_t } (\sigma_{min,t}^2 - 1) \right] \times \left\| \delta^{t,i}_{max}  \right\|^2 \\
&\ge - \rho^2_{max,t} \left[ \frac{1}{|\mathcal{S}_t|} (1 - \sigma_{min,t}^2 ) + 4 \epsilon (1 - \epsilon) \max_{i,j \in \mathcal{S}_t} \cos \left\langle \delta_{t,i}, \delta_{t,j} \right\rangle \right] \times \left\| \delta^{t,i}_{max}  \right\|^2 \\
&\ge - \rho^2_{max,t} \left[ \frac{1}{|\mathcal{S}_t|} (1 - \sigma_{min,t}^2 ) +  \max_{i,j \in \mathcal{S}_t} \cos \left\langle \delta_{t,i}, \delta_{t,j} \right\rangle \right] \times \left\| \delta^{t,i}_{max}  \right\|^2 
\end{align*} where last ineq because \(4 \epsilon (1 - \epsilon) < 1 \).
\end{proof}

\end{corollary}

\subsection{Theorem of LA-FedSGD-Corr}

\begin{theorem} (LA-FedSGD-Corr) 
\label{theorem:LA-FedSGD-Corr-App}
Assuming the clients satisfy Assumptions \ref{assumption:local_optimal_heter} and \ref{assumption:eff_dec}, and the updating model takes the form of \(x_{i,K}^{h,t} = x_t - \eta_l^i \sum_{i=0}^{K_i-1} \nabla f_i + \eta_l^i K_i h_t = x_{i,K}^t + h_t^i\). Let \(\delta_{\mathcal{S}_t}^K = x_{t+1} - x^*_{\mathcal{S}_t}\) and \(\delta_{\mathcal{S}_t} = x_t - x^*_{\mathcal{S}_t}\), the single-round update distance is:
\begin{equation}
\left\| x_{t+1} - x^*_{\mathcal{S}_t} \right\|^2  = 
(1 - \eta_t\wp) \left\| x_{t} - x^*_{\mathcal{S}_t} \right\|^2  - \eta_t\left( \eta_t\Delta_{t+1} + \eth \left\| \mathbb{H}_{\mathcal{S}_t}   \right\| \right),
\end{equation}  where \( \wp = 2(1 - \eta_t) \left( 1 - \cos \langle \delta_{\mathcal{S}_t}^K , \delta_{\mathcal{S}_t} \rangle \sigma_\Delta  \right) , \eth = [  2 \hbar \left\| \delta_{\mathcal{S}_t}  \right\| - \eta_t \left\| \mathbb{H}_{\mathcal{S}_t}  \right\| ] , \hbar =  - \eta_t\sigma_{\Delta} \cos \langle \delta_{\mathcal{S}_t}^K , \mathbb{H}_{\mathcal{S}_t}  \rangle + (\eta_t- 1) \cos \langle \delta_{\mathcal{S}_t} , \mathbb{H}_{\mathcal{S}_t}  \rangle  \) and \(\sigma_{\Delta_{t+1}} = \sqrt{\left\| \delta_{\mathcal{S}_t}  \right\| - \Delta_{t+1}}/\left\| \delta_{\mathcal{S}_t}  \right\|  , \mathbb{H}_{\mathcal{S}_t} = \sum_{i \in \mathcal{S}_t} \rho_i h_t^i \).
\begin{proof}

Through Lemma \ref{lemma:pseudo-gradient}, we can know that the pseudo-gradient and LA-FedAVG algorithm vector are consistent. For the starting operation of LA-FedSGD-Corr (referred to as LAFC for short), it involves substituting the pseudo-gradient using \( x_{t+1}^{\text{LA-FedAVG}} - x_t = -\mathbb{G}_{\mathcal{S}_t} \) to establish a connection with Theorem \ref{theorem:main}.

Assume client update as follow :
\begin{align*}
& x_{i,K}^{h,t} = x_{t} - \eta_l^i \sum_{i=0}^{K^i-1} \nabla f_i + \eta_l^i K^i h_t = x_{i,K}^{t} +  h_t^i  
\end{align*}

The update \(x_{t+1}^{\text{LAFC}}\) of \ref{alg:FedNAOMPX} is 
\begin{align*}
x_{t+1}^{\text{LAFC}} - x_t  &= - \eta_t\left[ \sum_{i \in \mathcal{S}_t} \rho_i (x_t - x_{i,K}^{t} -  h_t^i) \right]  = - \eta_t\left(\mathbb{G}_{\mathcal{S}_t} - \mathbb{H}_{\mathcal{S}_t} \right)
\end{align*} where \(  \mathbb{G}_{\mathcal{S}_t} = \sum_{i \in \mathcal{S}_t} \rho_i (x_t - x_{i,K}^{t}) \) and \( \mathbb{H}_{\mathcal{S}_t} = \sum_{i \in \mathcal{S}_t} \rho_i h_t^i \).

\begin{align*}
x_{t+1}^{\text{LAFC}} - x_t &= - \eta_t\left(\mathbb{G}_{\mathcal{S}_t} - \mathbb{H}_{\mathcal{S}_t} \right) \\
x_{t+1}^{\text{LAFC}} - x_{\mathcal{S}_t}^* - ( x_t - x_{\mathcal{S}_t}^*) &= \eta_t\left( x_{t+1}^{\text{LA-FedAVG}} - x_t \right) + \eta_t\mathbb{H}_{\mathcal{S}_t} \\
&= \eta_t\left( x_{t+1}^{\text{LA-FedAVG}} - x_{\mathcal{S}_t}^* -  (x_t - x_{\mathcal{S}_t}^*) \right) + \eta_t\mathbb{H}_{\mathcal{S}_t} \\
x_{t+1}^{\text{LAFC}} - x_{\mathcal{S}_t}^* &= \eta_t\left( x_{t+1}^{\text{LA-FedAVG}} - x_{\mathcal{S}_t}^* \right) + (1 -\eta_t) ( x_t - x_{\mathcal{S}_t}^*) + \eta_t\mathbb{H}_{\mathcal{S}_t} 
\end{align*} 

\begin{align*}
\left\| x_{t+1}^{\text{LAFC}} - x_{\mathcal{S}_t}^* \right\|^2 &= \left\| \eta_t\left( x_{t+1}^{\text{LA-FedAVG}} - x_{\mathcal{S}_t}^* \right) + (1 -\eta_t) ( x_t - x_{\mathcal{S}_t}^*) + \eta_t\mathbb{H}_{\mathcal{S}_t} \right\|^2 \\
&= \eta^2_t \left\| \delta_{\mathcal{S}_t}^K  \right\|^2 + (1 - \eta_t)^2 \left\| \delta_{\mathcal{S}_t}  \right\|^2 + \eta^2_t \left\| \mathbb{H}_{\mathcal{S}_t}  \right\|^2  \\
&+ 2 \eta_t(1 - \eta_t) \left\| \delta_{\mathcal{S}_t}^K  \right\| \left\| \delta_{\mathcal{S}_t}  \right\| \cos \langle \delta_{\mathcal{S}_t}^K , \delta_{\mathcal{S}_t} \rangle \\
&+ 2 \eta^2_t \left\| \delta_{\mathcal{S}_t}^K  \right\| \left\| \mathbb{H}_{\mathcal{S}_t}   \right\| \cos \langle \delta_{\mathcal{S}_t}^K , \mathbb{H}_{\mathcal{S}_t}  \rangle \\
&+ 2 \eta_t(1 - \eta_t) \left\| \delta_{\mathcal{S}_t}  \right\| \left\| \mathbb{H}_{\mathcal{S}_t}   \right\| \cos \langle \delta_{\mathcal{S}_t} , \mathbb{H}_{\mathcal{S}_t}  \rangle \\
&= \eta^2_t \left( \left\| \delta_{\mathcal{S}_t}  \right\|^2 - \Delta \right) + (1 - \eta_t)^2 \left\| \delta_{\mathcal{S}_t}  \right\|^2 + \eta^2_t \left\| \mathbb{H}_{\mathcal{S}_t}  \right\|^2  \\
&+ 2 \eta_t(1 - \eta_t) \sqrt{\left\| \delta_{\mathcal{S}_t}  \right\|^2 - \Delta } \left\| \delta_{\mathcal{S}_t}  \right\| \cos \langle \delta_{\mathcal{S}_t}^K , \delta_{\mathcal{S}_t} \rangle \\
&+ 2 \eta^2_t \sqrt{\left\| \delta_{\mathcal{S}_t}  \right\|^2 - \Delta } \left\| \mathbb{H}_{\mathcal{S}_t}   \right\| \cos \langle \delta_{\mathcal{S}_t}^K , \mathbb{H}_{\mathcal{S}_t}  \rangle \\
&+ 2 \eta_t(1 - \eta_t) \left\| \delta_{\mathcal{S}_t}  \right\| \left\| \mathbb{H}_{\mathcal{S}_t}   \right\| \cos \langle \delta_{\mathcal{S}_t} , \mathbb{H}_{\mathcal{S}_t}  \rangle \\
&= \left( 2 \eta^2_t - 2 \eta_t+ 1  \right) \left\| \delta_{\mathcal{S}_t}  \right\|^2 - \eta^2_t \Delta + \eta^2_t \left\| \mathbb{H}_{\mathcal{S}_t}  \right\|^2 \\
&+ 2 \eta_t(1 - \eta_t) \cos \langle \delta_{\mathcal{S}_t}^K , \delta_{\mathcal{S}_t} \rangle \sigma_\Delta \left\| \delta_{\mathcal{S}_t}  \right\|^2  \\
&+ \left[ 2 \eta^2_t \sigma_\Delta \cos \langle \delta_{\mathcal{S}_t}^K , \mathbb{H}_{\mathcal{S}_t}  \rangle + 2 \eta_t(1 - \eta_t) \cos \langle \delta_{\mathcal{S}_t} , \mathbb{H}_{\mathcal{S}_t}  \rangle \right] \left\| \delta_{\mathcal{S}_t}  \right\| \left\| \mathbb{H}_{\mathcal{S}_t}   \right\|  \\
&= \left( 2 \eta^2_t - 2 \eta_t+ 1 + 2 \eta_t(1 - \eta_t) \cos \langle \delta_{\mathcal{S}_t}^K , \delta_{\mathcal{S}_t} \rangle \sigma_\Delta  \right) \left\| \delta_{\mathcal{S}_t}  \right\|^2 - \eta^2_t \Delta + \eta^2_t \left\| \mathbb{H}_{\mathcal{S}_t}  \right\|^2 \\
&+ \left[ 2 \eta^2_t \sigma_\Delta \cos \langle \delta_{\mathcal{S}_t}^K , \mathbb{H}_{\mathcal{S}_t}  \rangle + 2 \eta_t(1 - \eta_t) \cos \langle \delta_{\mathcal{S}_t} , \mathbb{H}_{\mathcal{S}_t}  \rangle \right] \left\| \delta_{\mathcal{S}_t}  \right\| \left\| \mathbb{H}_{\mathcal{S}_t}   \right\|
\end{align*} where \(
\delta_{\mathcal{S}_t}^K =  x_{t+1}^{\text{LA-FedAVG}} - x_{\mathcal{S}_t}^* , \delta_{\mathcal{S}_t} = x_t - x_{\mathcal{S}_t}^* 
\)

\begin{align*}
& \left\| x_{t+1}^{\text{LAFC}} - x_{\mathcal{S}_t}^* \right\|^2 - \left\| \delta_{\mathcal{S}_t}  \right\|^2 \\ 
&= 2 \eta_t(\eta_t- 1) \left( 1 - \cos \langle \delta_{\mathcal{S}_t}^K , \delta_{\mathcal{S}_t} \rangle \sigma_\Delta  \right) \left\| \delta_{\mathcal{S}_t}  \right\|^2 - \eta^2_t \Delta + \eta^2_t \left\| \mathbb{H}_{\mathcal{S}_t}  \right\|^2 \\
&+ \left[ 2 \eta^2_t \sigma_\Delta \cos \langle \delta_{\mathcal{S}_t}^K , \mathbb{H}_{\mathcal{S}_t}  \rangle + 2 \eta_t(1 - \eta_t) \cos \langle \delta_{\mathcal{S}_t} , \mathbb{H}_{\mathcal{S}_t}  \rangle \right] \left\| \delta_{\mathcal{S}_t}  \right\| \left\| \mathbb{H}_{\mathcal{S}_t}  \right\|  \\
&= -\eta_t\wp \left\| \delta_{\mathcal{S}_t}  \right\|^2 - \eta^2_t \Delta_{t+1} - \eta_t \eth \left\| \mathbb{H}_{\mathcal{S}_t}   \right\| 
\end{align*} where \( \wp = 2(1 - \eta_t) \left( 1 - \cos \langle \delta_{\mathcal{S}_t}^K , \delta_{\mathcal{S}_t} \rangle \sigma_\Delta  \right) , \hbar = \left( - \eta_t\sigma_\Delta \cos \langle \delta_{\mathcal{S}_t}^K , \mathbb{H}_{\mathcal{S}_t}  \rangle + (\eta_t- 1) \cos \langle \delta_{\mathcal{S}_t} , \mathbb{H}_{\mathcal{S}_t}  \rangle \right) \), \(\eth = \left[  2 \hbar \left\| \delta_{\mathcal{S}_t}  \right\| - \eta_t \left\| \mathbb{H}_{\mathcal{S}_t}  \right\| \right] \).

The correct term \(\left\| \mathbb{H}_{\mathcal{S}_t}  \right\| \) is absolutely effective if \(\eth > 0 \) : 

\begin{align*}
&\eth = \left[  2 \hbar \left\| \delta_{\mathcal{S}_t}  \right\| - \eta_t \left\| \mathbb{H}_{\mathcal{S}_t}  \right\| \right] > 0 \Rightarrow 2 \frac{\hbar}{\eta_t} \left\| \delta_{\mathcal{S}_t}  \right\| >  \left\| \mathbb{H}_{\mathcal{S}_t}  \right\| \Rightarrow  \left\| \mathbb{H}_{\mathcal{S}_t} \right\| < 2 \frac{\hbar}{\eta_t} \left\| \delta_{\mathcal{S}_t}  \right\| \\
&\hbar =  - \eta_t\sigma_\Delta \cos \langle \delta_{\mathcal{S}_t}^K , \mathbb{H}_{\mathcal{S}_t}  \rangle + (\eta_t- 1) \cos \langle \delta_{\mathcal{S}_t} , \mathbb{H}_{\mathcal{S}_t}  \rangle > 0 \Rightarrow \frac{1 - \eta_t}{\eta_t\sigma_\Delta} \cos \langle \delta_{\mathcal{S}_t} , \mathbb{H}_{\mathcal{S}_t} \rangle <  \cos \langle \delta_{\mathcal{S}_t}^K , \mathbb{H}_{\mathcal{S}_t}  \rangle
\end{align*}
\end{proof}
\end{theorem}

\subsection{Theorem of LA-SA}

For LA-SA, we refer to the form of QHM ~\citep{gitman2019understanding}:
\(d^t = (1 - \beta_t) \mathbb{G}_{\mathcal{S}_t} + \beta_t d^{t-1}, x^{t+1} = x^t - \eta_t \left[ (1 - \nu_t) \mathbb{G}_{\mathcal{S}_t} + \nu_t d^t \right] \)
where the parameter $\nu_t \in [0, 1]$ interpolates between SGD ~\citep{robbins1951stochastic}($\nu_t = 0$) and (normalized) SHB\citep{polyak1964some} ($\nu_t = 1$). When the parameters $\eta_t$, $\beta_t$ and $\nu_t$ are held constant (thus the subscript $t$ can be omitted) and $\nu = \beta$, it recovers a normalized variant of NAG~\citep{nesterov2018} with an additional coefficient $1 - \beta_t$ on the stochastic gradient term. For adaptive learning rate methods (e.g., Adam\(\left(\phi(\mathbb{G}_{\mathcal{S}_t}) = \sqrt{\beta_2 \mathbb{G}_{\mathcal{S}_t}^2 + (1 - \beta_2) v_{t-1}} \right)\)~\citep{kingma2014adam}, RMSProp~\citep{hinton2012neural}, Adagrad~\citep{duchi2011adaptive}, etc.), we simply abstract them as \(\eta_t^\phi = \frac{\eta_t}{\phi(\mathbb{G}_{\mathcal{S}_t})}\).

\begin{theorem} (LA-SA) Assuming the clients satisfy Assumptions \ref{assumption:local_optimal_heter} and \ref{assumption:eff_dec} ,and the single-round update distance is: 
\begin{equation}
\left\| x_{t+1} - x^*_{\mathcal{S}_t} \right\|^2  = \left(1 - \wp_{\nu,\beta}^{\phi,t}  \right) \left\| \delta_{\mathcal{S}_t}^t \right\|^2  - \eta_t^\phi \left( \eta_t^\phi \left( 1 - \nu_t \beta_t \right) \Delta_{t+1} + \nu_t \beta_t  \eth_\phi \left\|  d^{t-1}  \right\| \right),
\end{equation}  where \( \wp_{\nu,\beta}^{\phi,t}  = 2 \hat{\eta}_{\nu,\beta}^{\phi,t} (1 - \hat{\eta}_{\nu,\beta}^{\phi,t}) ( 1 - \cos \langle \delta_{\mathcal{S}_t}^K , \delta_{\mathcal{S}_t} \rangle \sigma_\Delta  ) , \eth_\phi = [ \eta_{\nu,\beta}^{\phi,t} \left\|  d^{t-1} \right\| - 2 \hbar_{\phi} \left\| \delta_{\mathcal{S}_t}  \right\| ] , \hbar_\phi = ( (\hat{\eta}_{\nu,\beta}^{\phi,t} - 1 ) \cos \langle \delta_{\mathcal{S}_t} ,  d^{t-1} \rangle - \hat{\eta}_{\nu,\beta}^{\phi,t} \sigma_\Delta \cos \langle \delta_{\mathcal{S}_t}^K ,  d^{t-1} \rangle ) \) and \( \hat{\eta}_{\nu,\beta}^{\phi,t} =\eta_t^\phi \left( 1 - \nu_t \beta_t \right) , \eta_{\nu,\beta}^{\phi,t} = - \eta_t^\phi \nu_t \beta_t\).

\begin{proof}

Adaptive Method (\(h_t = 0\)):

\begin{align*}
x_{t+1}^{\text{LA-SA}} &= x_t - \frac{\eta_t}{\phi(\mathbb{G}_{\mathcal{S}_t})} ( (1 - \nu_t) \mathbb{G}_{\mathcal{S}_t} + \nu_t d^t ) \Rightarrow
x_{t+1}^{\text{LA-SA}} - x_t  = - \hat{\eta}_{\nu,\beta}^{\phi,t} \mathbb{G}_{\mathcal{S}_t}  + \eta_{\nu,\beta}^{\phi,t} d^{t-1} 
\end{align*} where \(d^t = (1 - \beta_t) \mathbb{G}_{\mathcal{S}_t} + \beta_t d^{t-1} ,  \hat{\eta}_{\nu,\beta}^{\phi,t} =\eta_t^\phi \left( 1 - \nu_t \beta_t \right) \text{~and~} \eta_{\nu,\beta}^{\phi,t} = - \eta_t^\phi \nu_t \beta_t\).

\begin{align*}
x_{t+1}^{\text{LA-SA}} - x_{\mathcal{S}_t}^* - ( x_t - x_{\mathcal{S}_t}^*) &= \hat{\eta}_{\nu,\beta}^{\phi,t} \left( x_{t+1}^{\text{LA-FedAVG}} - x_{\mathcal{S}_t}^* -  (x_t - x_{\mathcal{S}_t}^*) \right)  + \eta_{\nu,\beta}^{\phi,t} d^{t-1} \\
x_{t+1}^{\text{LA-SA}} - x_{\mathcal{S}_t}^* &= \hat{\eta}_{\nu,\beta}^{\phi,t} \left( x_{t+1}^{\text{LA-FedAVG}} - x_{\mathcal{S}_t}^* \right) + (1 - \hat{\eta}_{\nu,\beta}^{\phi,t})(x_t - x_{\mathcal{S}_t}^*) + \eta_{\nu,\beta}^{\phi,t} d^{t-1} 
\end{align*}

\begin{align*}
\left\| x_{t+1}^{\text{LA-SA}} - x_{\mathcal{S}_t}^* \right\|^2 &= \left\| \hat{\eta}_{\nu,\beta}^{\phi,t} \left( x_{t+1}^{\text{LA-FedAVG}} - x_{\mathcal{S}_t}^* \right) + (1 - \hat{\eta}_{\nu,\beta}^{\phi,t})(x_t - x_{\mathcal{S}_t}^*) + \eta_{\nu,\beta}^{\phi,t} d^{t-1}  \right\|^2 \\
&= (\hat{\eta}_{\nu,\beta}^{\phi,t})^2 \left\| \delta_{\mathcal{S}_t}^K  \right\|^2 + (1 - \hat{\eta}_{\nu,\beta}^{\phi,t})^2 \left\| \delta_{\mathcal{S}_t}  \right\|^2 + (\eta_{\nu,\beta}^{\phi,t})^2 \left\|  d^{t-1} \right\|^2  \\
&+ 2 \hat{\eta}_{\nu,\beta}^{\phi,t} (1 - \hat{\eta}_{\nu,\beta}^{\phi,t}) \left\| \delta_{\mathcal{S}_t}^K  \right\| \left\| \delta_{\mathcal{S}_t}  \right\| \cos \langle \delta_{\mathcal{S}_t}^K , \delta_{\mathcal{S}_t} \rangle \\
&+ 2 \hat{\eta}_{\nu,\beta}^{\phi,t} \eta_{\nu,\beta}^{\phi,t} \left\| \delta_{\mathcal{S}_t}^K  \right\| \left\|  d^{t-1}  \right\| \cos \langle \delta_{\mathcal{S}_t}^K ,  d^{t-1} \rangle \\
&+ 2 \eta_{\nu,\beta}^{\phi,t} (1 - \hat{\eta}_{\nu,\beta}^{\phi,t}) \left\| \delta_{\mathcal{S}_t}  \right\| \left\|  d^{t-1}  \right\| \cos \langle \delta_{\mathcal{S}_t} ,  d^{t-1} \rangle \\
&= (\hat{\eta}_{\nu,\beta}^{\phi,t})^2 \left( \left\| \delta_{\mathcal{S}_t}  \right\|^2 - \Delta \right) + (1 - \hat{\eta}_{\nu,\beta}^{\phi,t})^2 \left\| \delta_{\mathcal{S}_t}  \right\|^2 + (\eta_{\nu,\beta}^{\phi,t})^2 \left\|  d^{t-1} \right\|^2  \\
&+ 2 \hat{\eta}_{\nu,\beta}^{\phi,t} (1 - \hat{\eta}_{\nu,\beta}^{\phi,t}) \sqrt{\left\| \delta_{\mathcal{S}_t}  \right\|^2 - \Delta } \left\| \delta_{\mathcal{S}_t}  \right\| \cos \langle \delta_{\mathcal{S}_t}^K , \delta_{\mathcal{S}_t} \rangle \\
&+ 2 \hat{\eta}_{\nu,\beta}^{\phi,t} \eta_{\nu,\beta}^{\phi,t}\sqrt{\left\| \delta_{\mathcal{S}_t}  \right\|^2 - \Delta } \left\|  d^{t-1}  \right\| \cos \langle \delta_{\mathcal{S}_t}^K ,  d^{t-1} \rangle \\
&+ 2 \eta_{\nu,\beta}^{\phi,t} (1 - \hat{\eta}_{\nu,\beta}^{\phi,t}) \left\| \delta_{\mathcal{S}_t}  \right\| \left\|  d^{t-1}  \right\| \cos \langle \delta_{\mathcal{S}_t} ,  d^{t-1} \rangle \\
&= \left(2 (\hat{\eta}_{\nu,\beta}^{\phi,t})^2 - 2\hat{\eta}_{\nu,\beta}^{\phi,t}  + 1  + 2 \hat{\eta}_{\nu,\beta}^{\phi,t} (1 - \hat{\eta}_{\nu,\beta}^{\phi,t}) \cos \langle \delta_{\mathcal{S}_t}^K , \delta_{\mathcal{S}_t} \rangle \sigma_\Delta  \right) \left\| \delta_{\mathcal{S}_t}  \right\|^2 - (\hat{\eta}_{\nu,\beta}^{\phi,t})^2 \Delta \\
&+ (\eta_{\nu,\beta}^{\phi,t})^2 \left\|  d^{t-1} \right\|^2 \\
&+ \left[ 2 \hat{\eta}_{\nu,\beta}^{\phi,t} \eta_{\nu,\beta}^{\phi,t}\sigma_\Delta \cos \langle \delta_{\mathcal{S}_t}^K ,  d^{t-1} \rangle + 2 \eta_{\nu,\beta}^{\phi,t} (1 - \hat{\eta}_{\nu,\beta}^{\phi,t}) \cos \langle \delta_{\mathcal{S}_t} ,  d^{t-1} \rangle \right] \left\| \delta_{\mathcal{S}_t}  \right\| \left\|  d^{t-1}  \right\|  
\end{align*} where \(
\delta_{\mathcal{S}_t}^K =  x_{t+1}^{\text{LA-FedAVG}} - x_{\mathcal{S}_t}^* , \delta_{\mathcal{S}_t} = x_t - x_{\mathcal{S}_t}^* 
\)

\begin{align*}
& \left\| x_{t+1}^{\text{LA-SA}} - x_{\mathcal{S}_t}^* \right\|^2 - \left\| \delta_{\mathcal{S}_t}  \right\|^2 \\ 
&= \left(2 (\hat{\eta}_{\nu,\beta}^{\phi,t})^2 - 2\hat{\eta}_{\nu,\beta}^{\phi,t}  + 2 \hat{\eta}_{\nu,\beta}^{\phi,t} (1 - \hat{\eta}_{\nu,\beta}^{\phi,t}) \cos \langle \delta_{\mathcal{S}_t}^K , \delta_{\mathcal{S}_t} \rangle \sigma_\Delta  \right) \left\| \delta_{\mathcal{S}_t}  \right\|^2 - (\hat{\eta}_{\nu,\beta}^{\phi,t})^2 \Delta_{t+1} + (\eta_{\nu,\beta}^{\phi,t})^2 \left\|  d^{t-1} \right\|^2 \\
&+ \left[ 2 \hat{\eta}_{\nu,\beta}^{\phi,t} \eta_{\nu,\beta}^{\phi,t}\sigma_\Delta \cos \langle \delta_{\mathcal{S}_t}^K ,  d^{t-1} \rangle + 2 \eta_{\nu,\beta}^{\phi,t} (1 - \hat{\eta}_{\nu,\beta}^{\phi,t}) \cos \langle \delta_{\mathcal{S}_t} ,  d^{t-1} \rangle \right] \left\| \delta_{\mathcal{S}_t}  \right\| \left\|  d^{t-1}  \right\|  \\
&= -\hat{\eta}_{\nu,\beta}^{\phi,t} \wp \left\| \delta_{\mathcal{S}_t}  \right\|^2 - (\hat{\eta}_{\nu,\beta}^{\phi,t})^2 \Delta_{t+1}  \\
&+ \left[ (\eta_{\nu,\beta}^{\phi,t})^2 \left\|  d^{t-1} \right\| + \left(2 \hat{\eta}_{\nu,\beta}^{\phi,t} \eta_{\nu,\beta}^{\phi,t}\sigma_\Delta \cos \langle \delta_{\mathcal{S}_t}^K ,  d^{t-1} \rangle + 2 \eta_{\nu,\beta}^{\phi,t} (1 - \hat{\eta}_{\nu,\beta}^{\phi,t}) \cos \langle \delta_{\mathcal{S}_t} ,  d^{t-1} \rangle\right) \left\| \delta_{\mathcal{S}_t}  \right\| \right] \left\|  d^{t-1}  \right\| \\
&= -\hat{\eta}_{\nu,\beta}^{\phi,t} \wp \left\| \delta_{\mathcal{S}_t}  \right\|^2 - (\hat{\eta}_{\nu,\beta}^{\phi,t})^2 \Delta_{t+1}  \\
&+ \eta_{\nu,\beta}^{\phi,t} \left[ \eta_{\nu,\beta}^{\phi,t} \left\|  d^{t-1} \right\| + \left(2 \hat{\eta}_{\nu,\beta}^{\phi,t} \sigma_\Delta \cos \langle \delta_{\mathcal{S}_t}^K ,  d^{t-1} \rangle + 2(1 - \hat{\eta}_{\nu,\beta}^{\phi,t}) \cos \langle \delta_{\mathcal{S}_t} ,  d^{t-1} \rangle\right) \left\| \delta_{\mathcal{S}_t}  \right\| \right] \left\|  d^{t-1}  \right\| \\
&= -\hat{\eta}_{\nu,\beta}^{\phi,t} \wp \left\| \delta_{\mathcal{S}_t}  \right\|^2 - (\hat{\eta}_{\nu,\beta}^{\phi,t})^2 \Delta_{t+1}  \\
&- \eta_{\nu,\beta}^{\phi,t} \left[ -\eta_{\nu,\beta}^{\phi,t} \left\|  d^{t-1} \right\| + \left( - 2 \hat{\eta}_{\nu,\beta}^{\phi,t} \sigma_\Delta \cos \langle \delta_{\mathcal{S}_t}^K ,  d^{t-1} \rangle + 2(\hat{\eta}_{\nu,\beta}^{\phi,t} - 1 ) \cos \langle \delta_{\mathcal{S}_t} ,  d^{t-1} \rangle\right) \left\| \delta_{\mathcal{S}_t}  \right\| \right] \left\|  d^{t-1}  \right\| \\
&= - \eta_t^\phi \left( 1 - \nu_t \beta_t \right) \wp \left\| \delta_{\mathcal{S}_t}  \right\|^2 - (\eta_t^\phi \left( 1 - \nu_t \beta_t \right) )^2  \Delta_{t+1}  - \eta_t^\phi \nu_t \beta_t  \eth_\phi \left\|  d^{t-1}  \right\| 
\end{align*} where \( \wp = 2(1 - \hat{\eta}_{\nu,\beta}^{\phi,t}) \left( 1 - \cos \langle \delta_{\mathcal{S}_t}^K , \delta_{\mathcal{S}_t} \rangle \sigma_\Delta  \right) , \hbar_\phi =\\ ( - \hat{\eta}_{\nu,\beta}^{\phi,t} \sigma_\Delta \cos \langle \delta_{\mathcal{S}_t}^K ,  d^{t-1} \rangle + (\hat{\eta}_{\nu,\beta}^{\phi,t} - 1 ) \cos \langle \delta_{\mathcal{S}_t} ,  d^{t-1} \rangle ) , \eth_\phi = [ \eta_{\nu,\beta}^{\phi,t} \left\|  d^{t-1} \right\| - 2 \hbar_{\phi} \left\| \delta_{\mathcal{S}_t}  \right\| ]\).

And  \( \hat{\eta}_{\nu,\beta}^{\phi,t} =\eta_t^\phi \left( 1 - \nu_t \beta_t \right) \text{~and~} \eta_{\nu,\beta}^{\phi,t} = - \eta_t^\phi \nu_t \beta_t\).

The correct term \(\left\|  d^{t-1} \right\| \) is absolutely effective if \(\eth_\phi < 0 \) : 

\begin{align*}
&\eth = \left[  2 \hbar \left\| \delta_{\mathcal{S}_t}  \right\| - \eta_t \left\|  d^{t-1} \right\| \right] > 0 \Rightarrow 2 \frac{\hbar}{\eta_t} \left\| \delta_{\mathcal{S}_t}  \right\| >  \left\|  d^{t-1} \right\| \Rightarrow  \left\|  d^{t-1}\right\| < 2 \frac{\hbar}{\eta_t} \left\| \delta_{\mathcal{S}_t}  \right\| \\
&\hbar =  - \eta_t\sigma_\Delta \cos \langle \delta_{\mathcal{S}_t}^K ,  d^{t-1} \rangle + (\eta_t- 1) \cos \langle \delta_{\mathcal{S}_t} ,  d^{t-1} \rangle > 0 \\
&\Rightarrow \frac{1 - \eta_t}{\eta_t\sigma_\Delta} \cos \langle \delta_{\mathcal{S}_t} ,  d^{t-1}\rangle <  \cos \langle \delta_{\mathcal{S}_t}^K ,  d^{t-1} \rangle
\end{align*}
\end{proof}
\end{theorem}

\vskip 0.2in
\bibliography{references}
\bibliographystyle{unsrt}  






\end{document}